\documentclass[sigconf, screen, nonacm]{acmart}

\usepackage{booktabs}
\usepackage{amsmath,amsfonts,bm}
\usepackage{graphicx}
\usepackage{adjustbox}
\usepackage{siunitx}
\usepackage{multirow}
\usepackage{makecell}
\usepackage{subcaption}
\usepackage{float}
\definecolor{codepurple}{RGB}{102,45,145}
\usepackage{enumitem}
\setlist[itemize]{leftmargin=1.5em, topsep=2pt, itemsep=1pt}

\DeclareMathAlphabet{\mathsfit}{\encodingdefault}{\sfdefault}{m}{sl}
\SetMathAlphabet{\mathsfit}{bold}{\encodingdefault}{\sfdefault}{bx}{n}

\title{Mitigating Hallucinations via Inter-Layer Consistency Aggregation in Large Vision-Language Models}

\author{Kai Tang}
\authornote{Equal contribution.}
\affiliation{%
  \institution{Peking University}
  \country{}}

\author{Jinhao You}
\authornotemark[1]
\affiliation{\institution{University of Pennsylvania}\country{}}

\author{Yichen Guo}
\affiliation{\institution{Nanyang Technological University}\country{}}

\author{Yiding Sun}
\affiliation{\institution{Peking University}\country{}}

\author{Dongxu Zhang}
\affiliation{\institution{Tsinghua University}\country{}}

\author{Wenya Wang}
\affiliation{\institution{Nanyang Technological University}\country{}}

\author{Hanze Li}
\affiliation{\institution{Virginia Tech}\country{}}

\author{Tao Luo}
\affiliation{\institution{University of Electronic Science and Technology of China}\country{}}

\author{Renyuan Li}
\affiliation{\institution{University of Electronic Science and Technology of China}\country{}}

\author{Xiande Huang}
\authornote{Corresponding author.}
\affiliation{\institution{De Artificial Intelligence Lab}\country{}}

\author{Shanghang Zhang}
\authornotemark[2]
\affiliation{\institution{Peking University}\country{}}

\renewcommand{\shortauthors}{Tang et al.}


\begin{document}

%%
%% Abstract
%%
\begin{abstract}
Despite the impressive capabilities of Large Vision-Language Models (LVLMs), they remain susceptible to hallucinations, where generated content is inconsistent with the input image. Existing training-free hallucination mitigation methods often suffer from unstable performance and high sensitivity to hyperparameter settings, which limits their practicality and broader adoption. In this paper, we propose \textit{\textbf{D}ecoding with Inter-layer \textbf{C}onsistency via \textbf{L}ayer \textbf{A}ggregation} (DCLA), a training-free decoding mechanism that requires no retraining, fine-tuning, or access to external knowledge bases. Specifically, DCLA constructs a dynamic semantic reference by aggregating representations from previous layers and uses it to correct semantically deviated layers, thereby enforcing inter-layer consistency. Experiments across seven LVLMs and multiple benchmarks demonstrate the generality of DCLA: it surpasses standard decoding by 28.58 MME points on LLaVA1.5-7B and 42.6 MME points on Qwen2.5-VL, while improving POPE accuracy by 2.74 percentage points in the strongest setting. Code is available at \href{https://github.com/EasonAI-5589/LLaVA-Hallucination}{\textcolor{codepurple}{\texttt{https://github.com/EasonAI-5589/LLaVA-Hallucination}}}.

\end{abstract}

\maketitle
\begingroup
\renewcommand{\thefootnote}{}
\footnotetext{Contact email: \href{mailto:kaitang030113@gmail.com}{\textcolor{codepurple}{\texttt{kaitang030113@gmail.com}}}}
\footnotetext{Corresponding author: \href{mailto:shanghang@pku.edu.cn}{\textcolor{codepurple}{\texttt{shanghang@pku.edu.cn}}}}
\endgroup

\section{Introduction}
Large Vision-Language Models (LVLMs) have rapidly advanced in recent years, demonstrating impressive capabilities in aligning visual and textual modalities, which has notably enhanced their performance on multi-modal tasks such as visual question answering (VQA), image captioning, and complex geometric reasoning \cite{bai2023qwen,dai2023instructblip,liu2023visual,ye2024mplug,zhou2023infmllm,zhu2023minigpt,zhang2026pointcot}. Despite these advancements, LVLMs remain susceptible to hallucinations---generating syntactically plausible but visually ungrounded outputs \cite{liu2023mitigating,gunjal2024detecting,li2023evaluating,lovenia2023negative}. This issue severely compromises their reliability and limits their applicability in high-stakes fields such as medical report generation \cite{hartsock2024vision}, autonomous driving \cite{zhou2024vision}, and embodied AI systems \cite{ma2024survey}, where the accuracy and trustworthiness of generated text are crucial.

Recent studies have identified several causes of hallucinations in LVLMs, including over-reliance on statistical biases in training data such as object co-occurrence and background context \cite{li2023evaluating,chen2024multi,zhou2023analyzing}, the dominance of language priors over visual inputs during decoding \cite{guan2024hallusionbench,han2022visual,kaul2024throne}, and weak cross-modal attention in deeper layers that undermines visual-textual alignment \cite{an2024agla,yangunderstanding}. To address hallucinations in LVLMs, knowledge editing methods have been proposed \cite{jiang2024interpreting,chen2025attribution,khandelwal2024cross,zhou2023analyzing,perry2025dynamic}. These approaches typically aim to mitigate hallucinations by fine-tuning specific memory-related parameters within LVLMs \cite{jiang2024interpreting,chen2025attribution,khandelwal2024cross} or by injecting and revising factual information through external knowledge bases \cite{zhou2023analyzing,perry2025dynamic}. However, such methods generally treat hallucination as a static knowledge deficiency, overlooking the fact that information representations evolve dynamically across layers during inference.

Recently, several studies \cite{chuang2023dola,wangdamo,leng2024mitigating} have approached this problem from a training-free perspective. Wang et al.~\cite{wangdamo} observed that hallucinations in LVLMs tend to manifest as localized surges at the later layers, suppressing pre-existing and visually grounded information in the decoding distribution. Based on this observation, they proposed a training-free approach that mitigates hallucinations by injecting accumulated momentum into the information flow during inference. This design effectively suppresses the localized surges in the decoding distributions observed in later layers. However, their approach focused primarily on accumulating momentum across layers to guide the directional update of activations in later layers, without explicitly addressing the evolving semantic inconsistencies that occur across layers during inference. Consequently, factual information captured by earlier layers may still be attenuated or overridden by semantically divergent activations in later layers, ultimately leading to hallucinations. Additionally, their method is highly sensitive to hyperparameters.

\begin{figure*}[t]
  \centering
  \includegraphics[width=0.75\linewidth]{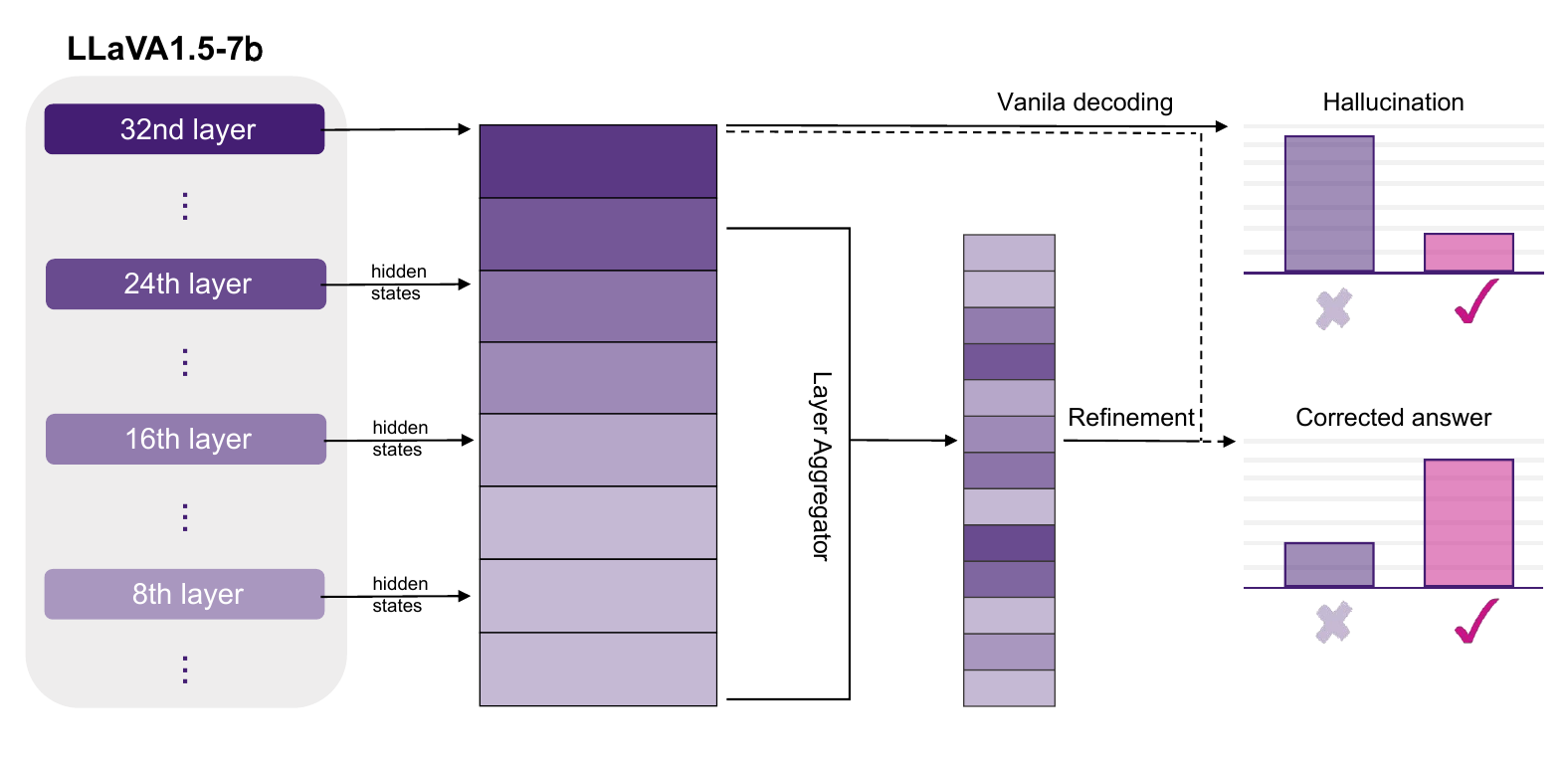}
  \caption{Illustration of Decoding with Inter-Layer Consistency via Layer Aggregation (DCLA) in LLaVA1.5-7b. Vanilla decoding process produces hallucinations and thus generates incorrect answers. By aggregating hidden states across layers to refine the representation, DCLA effectively suppresses hallucinations and restores the ground-truth answer.}
  \Description{A diagram showing how DCLA aggregates hidden states across transformer layers to correct hallucinated outputs in LLaVA1.5-7b, comparing vanilla decoding with DCLA-corrected decoding.}
  \label{fig:intro}
\end{figure*}

To address these challenges, we propose a novel decoding mechanism, \textbf{D}ecoding with Inter-layer \textbf{C}onsistency via \textbf{L}ayer \textbf{A}ggregation (DCLA), shown in Figure~\ref{fig:intro}. The key idea of DCLA is to treat the intermediate layer trajectory as an evolving semantic process rather than relying only on the final-layer representation. During decoding, DCLA maintains a dynamic semantic reference by aggregating hidden states from previous layers with distance-aware weights, so that visually grounded information captured earlier can remain available when later layers become dominated by language priors. For each layer, DCLA compares the current representation with this aggregated reference in the hidden-state space. When their semantic alignment falls below a threshold, the current hidden state is selectively refined through a lightweight fusion with the reference; otherwise, the model proceeds without intervention. This design differs from momentum-based correction in DAMO~\cite{wangdamo}, which mainly accumulates directional updates, as DCLA explicitly models inter-layer semantic consistency and applies corrections only when representation drift is detected. Since the refined hidden state is further used in subsequent aggregation, DCLA forms a self-stabilizing decoding trajectory that preserves reliable semantics while avoiding unnecessary perturbations.

Experiments on the MME and POPE benchmarks demonstrate that, without any additional training, DCLA significantly reduces hallucinations across four diverse LVLMs: LLaVA1.5-7b \cite{liu2023visual}, LLaVA1.5-13b \cite{liu2023visual}, LLaVA-NEXT \cite{liu2024llavanext}, and mPLUG-Owl2 \cite{ye2023mplug}. Results on VizWiz and MM-Vet datasets show the broader applicability of our method beyond hallucination mitigation. Experiments on Qwen2.5-VL, Qwen3-VL, and InternVL3 further confirm that DCLA generalizes across model families with minimal overhead. The contributions are summarized as follows:

\begin{itemize}
    \item We identify inter-layer semantic inconsistency as a key factor behind hallucinations in LVLM decoding, providing a dynamic perspective beyond static knowledge deficiency or final-layer output correction.
    \item We propose Decoding with Inter-layer Consistency via Layer Aggregation (DCLA), a training-free decoding method that constructs a layer-wise semantic reference and adaptively refines semantically deviated hidden states without retraining or external knowledge.
    \item We conduct extensive experiments across seven LVLMs and multiple benchmarks, demonstrating that DCLA consistently improves hallucination mitigation and general multimodal reliability with limited inference overhead.
\end{itemize}

\section{Related Work}
\paragraph{Large Vision-Language Models}
LVLMs have evolved from BERT-based architectures \cite{devlin2019bert,liu2019roberta,sun2019videobert,chen2020uniter,lu2019vilbert,tan2019lxmert}, which were designed to integrate visual and textual information, into a paradigm driven by large language models (LLMs) \cite{zhu2023minigpt,zhou2023infmllm,liu2023visual,ye2024mplug,bai2023qwen}. Supported by end-to-end training techniques \cite{jia2021scaling,radford2021learning}, recent LVLMs can perform unified decoding over visual and textual tokens, which substantially improves their expressiveness and adaptability. Models such as LLaVA \cite{liu2023visual} and InstructBLIP \cite{dai2023instructblip} further refine this paradigm through visual instruction tuning, enhancing performance across diverse vision-language tasks and highlighting the growing trend toward task-specific adaptation.

\paragraph{Layer Aggregation Mechanisms}
In the field of computer vision, Donahue et al.~\cite{donahue2014decaf} and Yosinski et al.~\cite{yosinski2014transferable} have pointed out that as the depth of neural networks increases, high-level representations gain stronger semantic abstraction capabilities, while fine-grained features such as edges and textures are often gradually forgotten. To address this, multi-level feature aggregation mechanisms have been proposed to enhance semantic representation and structural perception \cite{yu2018deep}. Building on this, the concept of layer aggregation has been widely adopted for efficient feature integration and mitigating the loss of shallow features \cite{huang2020dianet,zhao2021recurrence}. In the domain of large foundation models, the idea of layer aggregation has also been extensively adopted \cite{tenney2019bert,brandon2024reducing,wu2024layer,zhou2024value,li2025drafts}. To alleviate memory consumption and improve throughput during inference, Brandon et al.~\cite{brandon2024reducing}, Wu and Tu~\cite{wu2024layer}, and Zhou et al.~\cite{zhou2024value} have adopted layer aggregation strategies to reduce the size of key value caches, effectively accelerating model execution. Related studies on efficient reasoning further show that compression and adaptive refinement should preserve reliable intermediate semantics rather than treating all reasoning steps uniformly \cite{zhang2026chain,zhang2025not,zhang2026not}. Meanwhile, other studies \cite{tenney2019bert,li2025drafts} have explored the use of layer aggregation in the training or fine-tuning stages, demonstrating its effectiveness in enhancing the representation capacity and task-specific performance.

\paragraph{Hallucinations in LVLMs}
Hallucinations---generating content that deviates from the input or factual reality---are a well-studied problem in both LLMs \cite{lin2021truthfulqa,ji2023survey,zhu2024ibd,xu2024hallucination} and LVLMs \cite{li2023evaluating,liu2023mitigating,lovenia2023negative}, where tight visual-textual alignment makes them particularly challenging to mitigate \cite{yin2024woodpecker,li2025mitigating}. Training-based approaches address this through curated datasets \cite{wang2024mitigating,xiao2025detecting}, correction modules \cite{gunjal2024detecting,liu2023mitigating}, or improved cross-modal alignment \cite{kim2023exposing,zeng2021multi}, but they are often resource-intensive.

Training-free methods have attracted growing attention. DoLa \cite{chuang2023dola} mitigates hallucinations via contrastive decoding between layers. VCD \cite{leng2024mitigating} contrasts output distributions from original and distorted visual inputs. DAMO \cite{wangdamo} accumulates activations layer by layer as momentum to correct hidden states in deeper layers. Other approaches operate on attention or visual grounding: Opera \cite{huang2024opera} penalizes over-trust attention patterns, HALC \cite{chen2024halc} adopts focal-contrast decoding, Contrastive Region Guidance \cite{wan2024contrastive} contrasts model outputs with and without region-based visual prompts to improve grounding, Image-Grounded Guidance \cite{zhao2024mitigating} leverages external vision models to provide object-level evidence for constraining generation, Attentional Vision Calibration \cite{woo2025don} identifies over-attended irrelevant image tokens and applies contrastive decoding to counteract their influence, and Latent Space Steering \cite{liu2024reducing} intervenes in latent representations to stabilize vision features. These methods primarily operate on attention mechanisms, visual grounding, or output distributions. In contrast, DCLA targets a complementary dimension---enforcing semantic consistency across the transformer layer hierarchy during decoding.

\section{Method}

\begin{figure*}[t]
  \centering
  \includegraphics[width=0.75\linewidth]{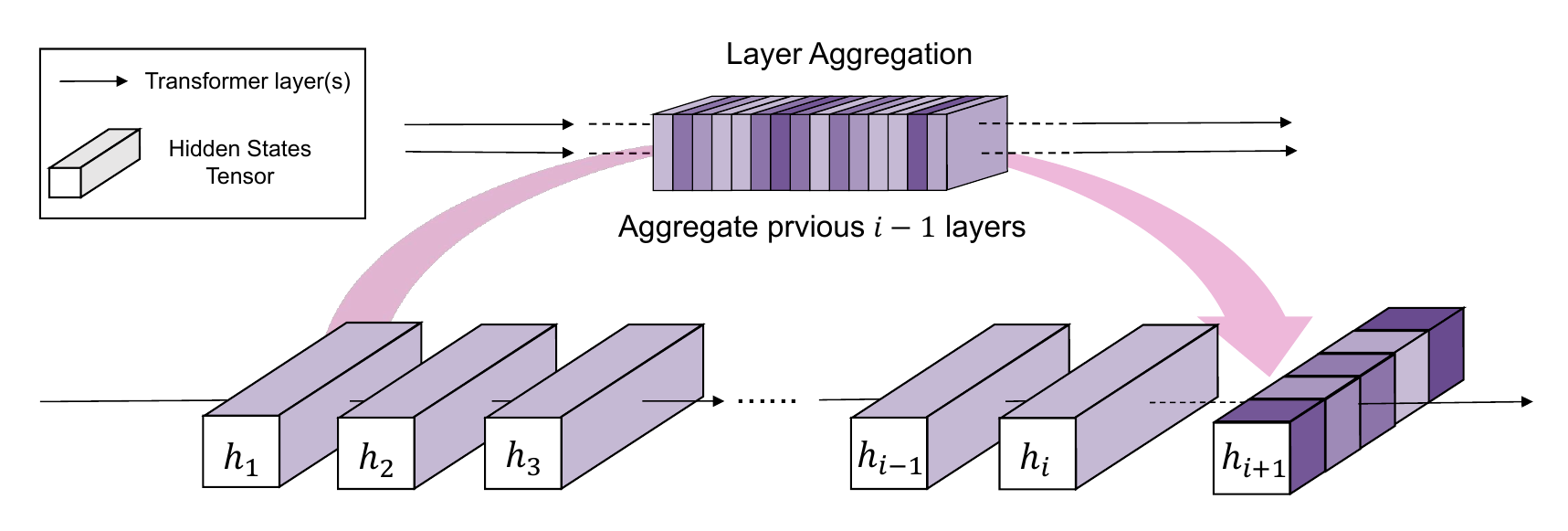}
  \caption{Layer Aggregation mechanism to form a stable semantic reference.}
  \Description{A schematic diagram showing the layer aggregation mechanism, where hidden states from earlier transformer layers are aggregated with exponential weighting to form a semantic reference for correcting later layers.}
  \label{fig:layerAgg}
\end{figure*}

\subsection{Preliminaries: Decoding in Large Vision-Language Models}
Large Vision-Language Models (LVLMs) generate textual output in an autoregressive manner through $N$ stacked transformer layers \cite{vaswani2017attention,liu2023visual,zhu2023minigpt}, where each layer plays a crucial role in the inference process. As information propagates forward through the network, feature representations are gradually transformed from low-level signals to high-level semantic representations \cite{rogers2021primer}. At time step $t$, given an initial fused multimodal representation \( x_t\), let $\mathbf{h}_t^{(i)} \in \mathbb{R}^{B \times L_t \times D}$ denote the hidden state at layer $i$, where $B$, $L_t$, and $D$ denote the batch size, sequence length, and hidden dimension, respectively. The forward process can be formulated as follows:
\begin{equation}
\mathbf{h}_t^{(i)} =
\begin{cases}
\text{Embedding}(x_t), & \text{if } i = 0 \\
\text{TransformerLayer}^{(i)}(\mathbf{h}_t^{(i-1)}), & \text{if } i = 1, \dots, N
\end{cases}
\end{equation}
where \( \text{TransformerLayer}^{(i)} \) represents the \( i \)-th transformer block, consisting of a multi-head self-attention mechanism and a feedforward neural network. The token prediction process, including both standard decoding from the final layer and optional early exit decoding from intermediate layers, can be generalized as follows:
\begin{equation}
p_i(x_{t+1} \mid x_{:t}) = \text{softmax}(\phi(\mathbf{h}_t^{(i)})), \quad i = 0,...,N
\end{equation}
where $\phi(\cdot)$ is the language modeling head that maps hidden states to vocabulary distributions. This unified formulation enables predictions from any layer, providing flexibility for efficient inference and layer-wise interpretability. Notably, the case $i = N$ corresponds to standard decoding using the final-layer hidden state, while $i< N$ indicates early exit from an intermediate layer.

\subsection{Motivation}
Recent work \cite{wangdamo} utilized the early exit mechanism to trace the evolution of token probabilities across the multi-layer inference process in LVLMs. Their findings indicate that hallucinations often manifest as localized surges at the later layers, which tend to override earlier, visually grounded information. To address this, they introduced a momentum-based correction method that aggregates activations across layers and utilizes the accumulated momentum to adjust the hidden states in the later layers.

While effective to some extent, this correction mechanism is highly sensitive to hyperparameters and does not directly address the fundamental issue: the inconsistency between layers. We argue that this inter-layer inconsistency disrupts the natural information flow during decoding. Specifically, earlier layers, though encoding simpler semantics, are more factually grounded in visual input, whereas later layers construct complex semantics but often undergo semantic drift that triggers hallucinations \cite{chuang2023dola}.

Motivated by this, we propose the \textbf{DCLA} mechanism, which enhances inter-layer consistency and maintains the flow of reliable semantics by aggregating earlier-layer representations to stabilize semantics throughout the decoding process. This approach effectively mitigates hallucinations caused by semantic drift, thereby improving both the factual accuracy and the robustness of the generated output.

\subsection{Decoding with Inter-Layer Consistency via Layer Aggregation}
\paragraph{Layer Aggregation}
In order to enforce inter-layer consistency, it is crucial to establish a stable semantic reference throughout the decoding process. Earlier layers encode basic semantic structures that remain robust to overfitting and noise introduced in deeper layers. Therefore, the use of historical representation can effectively anchor the inference trajectory of models, promote consistency across layers, and mitigate hallucinations. We perform weighted aggregation over all earlier layers to ensure that the semantic reference comprehensively integrates the complete semantic information. Formally, at the $i$-th layer, the aggregated representation is defined as:
\begin{equation}
H_{\text{agg}}^{(i)} = \sum_{j=0}^{i-1} \tilde{w}_j^{(i)} \cdot \tilde{h}_j,
\quad
\text{where} \quad
\tilde{h}_j =
\begin{cases}
\hat{h}_j, & \text{if } j \in \mathcal{C} \\
h_j, & \text{otherwise}
\end{cases}
\end{equation}
Here, $\tilde{h}_j$ denotes the effective hidden state of the $j$-th layer, and $\mathcal{C} \subseteq \{0, 1, \dots, i-1\}$ records all previously corrected layers. If layer $j$ has been corrected, its refined state $\hat{h}_j$ is used in place of the original $h_j$. This ensures that once a layer is corrected, the updated representation persistently contributes to subsequent aggregation and decoding. To prioritize layers that are closer to the current decoding step while still leveraging the semantic stability provided by earlier layers, we introduce a normalized exponential weighting scheme. This design reflects the intuition that recent layers contain more task-specific contextual refinements, whereas earlier layers offer foundational but potentially less context-aware semantics. The weight assigned to each layer $j$ is defined as:
\begin{equation}
\tilde{w}_j^{(i)} =
\frac{\exp\left(s(j, i)\right)}
{\sum\limits_{k=0}^{i-1} \exp\left(s(k, i)\right)}
\end{equation}
where $s(j, i)= j - (i-1)$ captures the relative distance between layer $j$ and the current layer $i$. Since $s(j,i) \leq 0$ and increases as $j$ approaches $i$, the exponential naturally assigns higher weights to layers closer to the current layer. The aggregation can also be viewed as a recurrent semantic memory:
\begin{equation}
\mathcal{M}_{i-1}=\{\tilde{h}_0,\tilde{h}_1,\dots,\tilde{h}_{i-1}\}, \quad
H_{\text{agg}}^{(i)} = A(\mathcal{M}_{i-1}),
\end{equation}
where $A(\cdot)$ denotes the normalized exponential aggregator. Compared to linear or uniform weighting strategies, this approach provides a more flexible and natural decay pattern, which aligns with the hierarchical nature of transformer representations where semantic granularity deepens progressively across layers.
\begin{table*}[t]
  \centering
  \caption{Perception performance of LLaVA1.5-7b on the MME dataset with different correction ranges. Each column indicates a hallucination category, and \textbf{bold} values denote the best results.}
  \label{tab:perception_layers}
  \resizebox{\textwidth}{!}{
  \begin{tabular}{lccccccccccc}
    \toprule
    \textbf{Correction} &
    \textbf{Exist.} & \textbf{Count} & \textbf{Pos.} & \textbf{Color} & \textbf{Post.} & \textbf{Celeb.} & \textbf{Scene} & \textbf{Land.} & \textbf{Art.} & \textbf{OCR} & \textbf{Total} \\
    \midrule
    baseline       & 190.00 & \textbf{160.00} & 138.33 & 165.00 & 140.48 & 135.00 & 156.25 & 162.25 & 119.25 & 125.00 & 1491.56 \\
    1--16 layers   & \textbf{190.00} & 145.00 & \textbf{138.33} & \textbf{180.00} & 141.16 & \textbf{132.94} & \textbf{161.75} & 165.50 & \textbf{122.25} & 140.00 & \textbf{1516.93} \\
    1--20 layers   & 190.00 & 145.00 & 138.33 & 175.00 & 141.16 & 132.06 & 160.25 & 167.00 & 119.25 & \textbf{147.50} & 1515.55 \\
    1--24 layers   & 190.00 & 150.00 & 133.33 & 170.00 & 142.18 & 132.94 & 161.00 & 167.00 & 120.00 & 145.00 & 1511.45 \\
    1--28 layers   & 190.00 & 140.00 & 138.33 & 180.00 & 141.16 & 135.88 & 158.75 & 163.75 & 119.50 & 147.50 & 1514.87 \\
    1--32 layers   & 185.00 & 120.00 & 108.33 & 160.00 & \textbf{155.78} & 135.88 & 160.00 & 161.75 & 109.75 & 107.50 & 1404.00 \\
    \bottomrule
  \end{tabular}
  }
\end{table*}

\paragraph{Decoding with Inter-Layer Consistency}
In standard LVLMs, the decoding process typically follows the vanilla decoding strategy, where each transformer layer updates its hidden state solely based on the output of the preceding layer:
\begin{equation}
h_i^{\mathrm{std}}=\mathcal{T}_i\left(h_{i-1}^{\mathrm{std}}\right), \quad i=1,\dots,N,
\end{equation}
where $\mathcal{T}_i(\cdot)$ denotes the transformation performed by the $i$-th transformer layer. In this standard forward propagation process, information is passed through adjacent transformer blocks, while earlier hidden states are not explicitly reused after being transformed by subsequent layers. However, this unidirectional information flow can lead to the gradual loss of fine-grained semantic cues captured by earlier layers \cite{chuang2023dola,wangdamo}. To address this issue, we integrate the aggregated representation $H_{\text{agg}}^{(i)}$ into the decoding process. The hidden state is corrected through a linear fusion mechanism.
\begin{equation}
\hat{h}_i = \alpha \cdot h_i + (1 - \alpha) \cdot H_{\text{agg}}^{(i)}
\end{equation}
where $\alpha$ denotes the fusion coefficient between the original representation and the aggregated semantic reference. As shown in Figure~\ref{fig:layerAgg}, this correction mechanism allows the current layer to benefit from both localized contextual information and global aggregated semantics. Importantly, the corrected representation is not only used for the current layer output but is also stored as the effective state for subsequent aggregation. Thus, once a semantic deviation is corrected, the refined state becomes part of $\mathcal{M}_i$ and can guide later layers.

\subsection{Adaptive Layer Aggregation Correction}
Tenney et al.~\cite{tenney2019bert} have shown that different layers of the model perform different roles in semantic construction. Earlier layers focus more on capturing lower-level information, while later layers progressively inject abstract semantics and factual knowledge. Therefore, applying indiscriminate corrections to all layers may disrupt the natural evolution of information within the model.

To verify this hypothesis, we conducted preliminary experiments on the MME dataset, and the results in Table~\ref{tab:perception_layers} revealed that correcting different layers led to significant fluctuations in overall model performance. Moreover, increasing the number of corrected layers did not necessarily result in better outcomes. These observations suggest that correction should not be applied indiscriminately but rather selectively based on the intrinsic characteristics of each layer.

Thus, we propose a dynamic layer correction mechanism. During inference, the model adaptively determines whether a hallucination surge has occurred at each layer by comparing the semantic features of the current hidden states with the global aggregated vector. The dynamic layer selection mechanism, together with our layer aggregation strategy, completes our overall approach, Decoding with Inter-layer Consistency via Layer Aggregation (DCLA). To more directly capture the intrinsic semantic changes, we focus on the hidden states themselves rather than their derived probability distributions. Specifically, we do not employ an early exit mechanism to obtain intermediate output distributions. Instead, for each sample $b$ in a batch, we flatten the token and feature dimensions of $h_i^{(b)}$ and $H_{\text{agg}}^{(i,b)}$ using $\operatorname{vec}(\cdot)$, and compute a batch-level consistency score:
\begin{equation}
c_i =
\frac{1}{B}\sum_{b=1}^{B}
\frac{
\operatorname{vec}(h_i^{(b)})^\top \operatorname{vec}(H_{\text{agg}}^{(i,b)})
}{
\left\|\operatorname{vec}(h_i^{(b)})\right\|_2
\left\|\operatorname{vec}(H_{\text{agg}}^{(i,b)})\right\|_2
}.
\end{equation}
We adopt cosine similarity as the consistency metric because DCLA operates in the hidden state, where semantic drift manifests as \emph{directional} deviation between the current hidden state and the aggregated reference. Cosine similarity directly measures this directional alignment. In contrast, distribution-level divergences such as KL divergence operate on probability distributions and can only capture differences at the vocabulary level, failing to reflect directional shifts in the hidden state. Moreover, computing KL divergence at intermediate layers requires the language modeling head to project hidden states into distributions. Since this head is trained only on final-layer representations, its outputs may not accurately reflect the semantics of earlier layers.

Given a consistency threshold $\tau$, DCLA applies the correction only when $c_i < \tau$, which indicates that the current hidden state has deviated from the aggregated semantic reference. Otherwise, the model follows the original decoding trajectory without modifying the hidden state. Let $\tilde{h}_i$ denote the effective hidden state after this optional correction. The recurrent update of the semantic memory is then written as
\begin{equation}
\mathcal{M}_{i} = \mathcal{M}_{i-1} \cup \{\tilde{h}_i\}, \quad
h_{i+1} = \text{TransformerLayer}^{(i+1)}(\tilde{h}_i),
\end{equation}
which ensures that later layers are conditioned on the corrected state whenever semantic drift is detected. This adaptive strategy applies corrections only when necessary, preserving the model's natural semantic progression while mitigating potential inconsistencies.

\section{Experiment}

\begin{table*}[t]
\centering
\caption{Experimental results of various decoding strategies on MME dataset across four models: LLaVA1.5-7b, LLaVA-NEXT, LLaVA1.5-13b, and mPLUG-Owl2.}
\label{tab:mme_main_zeros}
\resizebox{\textwidth}{!}{
\begin{tabular}{llcccccccccccc}
\toprule
\textbf{Model} & \textbf{Decoding} & \textbf{Existence} & \textbf{Count} & \textbf{Position} & \textbf{Color} & \textbf{Posters} & \textbf{Celebrity} & \textbf{Scene} & \textbf{Landmark} & \textbf{Artwork} & \textbf{OCR} & \textbf{Total} \\
\midrule
\multirow{5}{*}{LLaVA1.5-7b}
& Regular & 190.00 & 160.00 & 138.33 & 165.00 & 140.48 & 135.00 & 156.25 & 162.25 & 119.25 & 125.00 & 1491.56  \\
& VCD     & 190.00 & 163.33 & 133.33 & 158.33 & 129.59 & \textbf{139.12} & 155.75 & \textbf{166.50} & \textbf{124.00} & 125.00 & 1484.96 \\
& DoLa    & 190.00 & 153.33 & 143.33 & 165.00 & 141.50 & 132.35 & 157.75 & 160.50 & 118.75 & 132.50 & 1495.02 \\
& DAMO    & \textbf{195.00} & 150.00 & 143.33 & 165.00 & \textbf{144.56} & 134.12 & \textbf{157.75} & 163.75 & 120.00 & 140.00 & 1513.51 \\
& \textbf{DCLA}    & 190.00 & \textbf{163.33} & \textbf{148.33} & \textbf{175.00} & 137.41 & 132.06 & 156.25 & 160.50 & 117.25 & \textbf{140.00} & \textbf{1520.14} \\
\midrule
\multirow{5}{*}{LLaVA-NEXT}
& Regular & 195.00 & 135.00 & 143.33 & 170.00 & 159.52 & 142.94 & \textbf{162.25} & 155.75 & 123.00 & 132.50 & 1519.30 \\
& VCD     & 175.00 & 125.00 & 95.00 & 140.00 & 148.98 & 145.29 & 159.00 & \textbf{169.75} & 130.25 & 130.00 & 1418.27 \\
& DoLa    & 190.00 & 133.33 & 143.33 & 170.00 & 132.65 & \textbf{155.59} & 156.25 & 135.00 & \textbf{136.75} & \textbf{162.50} & 1515.41 \\
& DAMO    & 195.00 & 130.00 & 133.33 & 160.00 & 149.32 & 145.29 & 159.50 & 143.25 & 123.75 & 132.50 & 1471.95 \\
& \textbf{DCLA}    & \textbf{195.00} & \textbf{140.00} & \textbf{143.33} & \textbf{170.00} & \textbf{160.20} & 142.94 & 161.50 & 155.75 & 124.50 & 132.50 & \textbf{1525.73} \\
\midrule
\multirow{5}{*}{LLaVA1.5-13b}
& Regular & 188.33 & 145.00 & 123.33 & 160.00 & 159.52 & 159.71 & 157.25 & 141.75 & 121.75 & 147.50 & 1504.15 \\
& VCD     & 190.00 & \textbf{163.33} & 120.00 & \textbf{175.00} & 151.70 & 159.41 & 158.25 & 129.00 & \textbf{125.25} & 132.50 & 1504.44 \\
& DoLa    & 190.00 & 150.00 & 123.33 & 160.00 & 160.54 & 157.06 & 155.75 & 134.50 & 124.00 & 147.50 & 1502.69 \\
& DAMO    & \textbf{190.00} & 125.00 & 113.33 & 150.00 & \textbf{166.66} & 152.06 & \textbf{163.25} & \textbf{166.50} & 107.75 & 140.00 & 1474.56 \\
& \textbf{DCLA}    & 188.33 & 145.00 & \textbf{123.33} & 160.00 & 159.52 & \textbf{160.88} & 158.00 & 140.50 & 121.75 & \textbf{147.50} & \textbf{1504.82} \\
\midrule
\multirow{5}{*}{mPLUG-Owl2}
& Regular & 185.00 & 165.00 & 78.33 & 150.00 & \textbf{163.27} & 162.94 & 152.75 & 160.00 & 139.75 & 102.50 & 1459.54 \\
& VCD     & 180.00 & 160.00 & 61.67 & 151.67 & 151.36 & 108.82 & 158.00 & 115.00 & 130.00 & 95.00 & 1311.52 \\
& DoLa    & 175.00 & 160.00 & \textbf{93.33} & \textbf{163.33} & 155.78 & 160.88 & \textbf{159.25} & 142.50 & \textbf{142.25} & \textbf{110.00} & 1462.33 \\
& DAMO    & 180.00 & 155.00 & 78.33 & 145.00 & 134.01 & \textbf{167.35} & 154.00 & \textbf{168.25} & 133.50 & 87.50 & 1402.95 \\
& \textbf{DCLA}    & \textbf{185.00} & \textbf{165.00} & 78.33 & 155.00 & 162.24 & 163.82 & 153.00 & 159.50 & 139.00 & 102.50 & \textbf{1463.40} \\
\bottomrule
\end{tabular}
}
\end{table*}

\begin{table*}[p]
\centering
\caption{Experimental results of various decoding strategies on the SEEM-annotated MSCOCO, A-OKVQA and GQA datasets from POPE using four models: LLaVA1.5-7b, LLaVA-NEXT, LLaVA1.5-13b, and mPLUG-Owl2. The best values for each metric across all models and decoding strategies are highlighted in \textbf{bold}.}
\label{tab:POPE_results}
\fontsize{8pt}{8.8pt}\selectfont
\renewcommand{\arraystretch}{0.9}
\setlength{\tabcolsep}{3.5pt}
\begin{tabular}{ll l cc cc cc}
\toprule
\multirow{2}{*}{\textbf{Setting}} & \multirow{2}{*}{\textbf{Model}} & \multirow{2}{*}{\textbf{Decoding}} & \multicolumn{2}{c}{\textbf{MSCOCO}} & \multicolumn{2}{c}{\textbf{A-OKVQA}} & \multicolumn{2}{c}{\textbf{GQA}} \\
\cmidrule(lr){4-5} \cmidrule(lr){6-7} \cmidrule(lr){8-9}
& & & Accuracy & F1 Score & Accuracy & F1 Score & Accuracy & F1 Score \\
\midrule
\multirow{20}{*}{Random}
& \multirow{5}{*}{LLaVA1.5-7b}
& Regular & 89.60 & 89.72 & 87.23 & 88.44 & 86.87 & 88.14 \\
& & VCD    & 89.07 & 89.00 & 86.37 & 87.44 & 86.00 & 87.33 \\
& & DoLa   & 89.70 & 89.79 & 86.10 & 87.48 & 85.47 & 87.00 \\
& & DAMO   & 89.94 & 89.90 & 87.83 & 88.88 & 86.03 & 87.52 \\
& & DCLA   & \textbf{90.03} & \textbf{89.99} & \textbf{87.93} & \textbf{88.98} & \textbf{87.90} & \textbf{88.94} \\
\cmidrule{2-9}
& \multirow{5}{*}{LLaVA-NEXT}
& Regular & 88.83 & 87.58 & 91.07 & 90.87 & 90.03 & 89.34 \\
& & VCD    & 80.00 & 75.14 & 81.13 & 77.45 & 81.13 & 77.29 \\
& & DoLa   & 85.40 & 83.02 & 88.73 & 87.68 & 86.97 & 85.34 \\
& & DAMO   & \textbf{88.90} & \textbf{87.69} & 90.87 & 90.43 & 88.63 & 87.86 \\
& & DCLA   & 88.87 & 87.60 & \textbf{91.10} & \textbf{90.69} & \textbf{90.10} & \textbf{89.52} \\
\cmidrule{2-9}
& \multirow{5}{*}{LLaVA1.5-13b}
& Regular & 88.37 & 87.15 & 91.03 & 90.72 & 91.03 & 90.75 \\
& & VCD    & 87.10 & 85.50 & 89.00 & 88.33 & 89.43 & 88.90 \\
& & DoLa   & 88.30 & 87.04 & 90.77 & 90.37 & \textbf{91.03} & 90.67 \\
& & DAMO   & 90.03 & 89.59 & \textbf{91.60} & \textbf{91.78} & 90.90 & \textbf{91.09} \\
& & DCLA   & \textbf{91.07} & \textbf{90.83} & 91.50 & 91.28 & 91.00 & 90.70 \\
\cmidrule{2-9}
& \multirow{5}{*}{mPLUG-Owl2}
& Regular & 88.40 & 87.71 & 88.09 & 88.17 & 86.10 & 85.41 \\
& & VCD    & 82.17 & 79.96 & 82.83 & 81.49 & 81.73 & 80.25 \\
& & DoLa   & 86.97 & 85.60 & 87.63 & 86.96 & 84.77 & 83.06 \\
& & DAMO   & \textbf{88.63} & \textbf{88.19} & \textbf{88.26} & \textbf{88.43} & 86.70 & 86.23 \\
& & DCLA   & 88.53 & 87.99 & 88.10 & 88.16 & \textbf{86.80} & \textbf{86.25} \\
\midrule
\multirow{20}{*}{Popular}
& \multirow{5}{*}{LLaVA1.5-7b}
& Regular & 86.20 & 86.81 & 80.10 & 83.07 & 74.50 & 79.29 \\
& & VCD    & 85.63 & 86.03 & 78.90 & 81.82 & 73.73 & 78.60 \\
& & DoLa   & 86.07 & 86.67 & 80.40 & 83.21 & 75.30 & 79.75 \\
& & DAMO   & 86.67 & 87.06 & 81.07 & 83.70 & \textbf{76.17} & \textbf{80.29} \\
& & DCLA   & \textbf{86.73} & \textbf{87.10} & \textbf{81.13} & \textbf{83.78} & 75.20 & 79.68 \\
\cmidrule{2-9}
& \multirow{5}{*}{LLaVA-NEXT}
& Regular & 87.63 & 86.43 & 89.13 & 88.87 & 86.57 & 86.23 \\
& & VCD    & 79.83 & 74.99 & 80.90 & 77.23 & 79.20 & 75.53 \\
& & DoLa   & 85.03 & 82.67 & 87.83 & 86.83 & 85.07 & 83.55 \\
& & DAMO   & 87.70 & \textbf{86.54} & 89.03 & 88.72 & 85.97 & 85.43 \\
& & DCLA   & \textbf{87.73} & 86.51 & \textbf{89.23} & \textbf{88.96} & \textbf{86.57} & \textbf{86.30} \\
\cmidrule{2-9}
& \multirow{5}{*}{LLaVA1.5-13b}
& Regular & 87.53 & 86.36 & 89.13 & 88.97 & 88.43 & 88.38 \\
& & VCD    & 86.13 & 84.58 & 86.73 & 86.26 & 86.53 & 86.28 \\
& & DoLa   & 87.53 & 86.31 & 88.93 & 88.68 & \textbf{88.60} & \textbf{88.43} \\
& & DAMO   & 88.87 & 88.51 & 89.07 & \textbf{89.56} & 86.77 & 87.55 \\
& & DCLA   & \textbf{90.27} & \textbf{90.09} & \textbf{89.17} & 89.14 & 88.40 & 88.33 \\
\cmidrule{2-9}
& \multirow{5}{*}{mPLUG-Owl2}
& Regular & 86.56 & 86.04 & 84.33 & 84.18 & 79.37 & 80.38 \\
& & VCD    & 80.13 & 77.89 & 80.77 & 79.93 & 77.83 & 77.12 \\
& & DoLa   & 85.77 & 84.48 & 84.47 & 84.15 & 81.37 & 80.29 \\
& & DAMO   & 86.50 & 86.28 & 84.07 & 84.92 & \textbf{79.93} & \textbf{80.59} \\
& & DCLA   & \textbf{86.83} & \textbf{86.45} & \textbf{84.43} & \textbf{85.06} & 79.87 & 80.44 \\
\midrule
\multirow{20}{*}{Adversarial}
& \multirow{5}{*}{LLaVA1.5-7b}
& Regular & 79.77 & 81.78 & 69.20 & 76.02 & 68.33 & 75.50 \\
& & VCD    & 79.27 & 81.01 & 68.30 & 74.97 & 68.10 & 75.15 \\
& & DoLa   & 79.47 & 81.52 & 69.60 & 76.16 & 68.90 & 75.77 \\
& & DAMO   & 80.50 & 82.14 & \textbf{70.64} & \textbf{76.85} & \textbf{70.07} & \textbf{76.43} \\
& & DCLA   & \textbf{80.70} & \textbf{82.28} & 70.23 & 76.60 & 69.13 & 75.91 \\
\cmidrule{2-9}
& \multirow{5}{*}{LLaVA-NEXT}
& Regular & 86.40 & 85.27 & 82.03 & 82.85 & 82.97 & 83.27 \\
& & VCD    & 79.13 & 74.34 & 75.70 & 72.72 & 76.23 & 72.98 \\
& & DoLa   & 84.30 & 81.97 & 82.83 & 82.37 & 81.87 & 80.71 \\
& & DAMO   & 86.33 & 85.26 & \textbf{83.16} & \textbf{83.68} & 82.53 & 82.49 \\
& & DCLA   & \textbf{86.53} & \textbf{85.38} & 82.33 & 83.08 & \textbf{83.17} & \textbf{83.40} \\
\cmidrule{2-9}
& \multirow{5}{*}{LLaVA1.5-13b}
& Regular & 85.63 & 84.60 & 81.93 & 82.91 & 83.73 & \textbf{84.40} \\
& & VCD    & 84.23 & 82.83 & 80.40 & 80.95 & 82.43 & 82.81 \\
& & DoLa   & 85.70 & 84.61 & 82.07 & 82.86 & 82.97 & 81.89 \\
& & DAMO   & 85.53 & 85.56 & 81.13 & \textbf{83.25} & 81.47 & 83.39 \\
& & DCLA   & \textbf{85.77} & \textbf{86.14} & \textbf{82.07} & 83.22 & \textbf{83.73} & 84.37 \\
\cmidrule{2-9}
& \multirow{5}{*}{mPLUG-Owl2}
& Regular & 84.20 & 83.98 & 76.67 & 78.71 & 78.07 & 79.03 \\
& & VCD    & 77.73 & 76.21 & 73.50 & 74.26 & 74.43 & 74.53 \\
& & DoLa   & 84.00 & 82.88 & \textbf{78.17} & 79.14 & 78.07 & 77.80 \\
& & DAMO   & 83.33 & 83.59 & 76.20 & 79.04 & 78.03 & 79.13 \\
& & DCLA   & \textbf{84.20} & \textbf{84.14} & 77.07 & \textbf{79.44} & \textbf{78.17} & \textbf{79.13} \\
\bottomrule
\end{tabular}
\end{table*}

\begin{figure*}[t]
    \centering
    \begin{subfigure}[t]{0.40\textwidth}
        \centering
        \includegraphics[width=\textwidth]{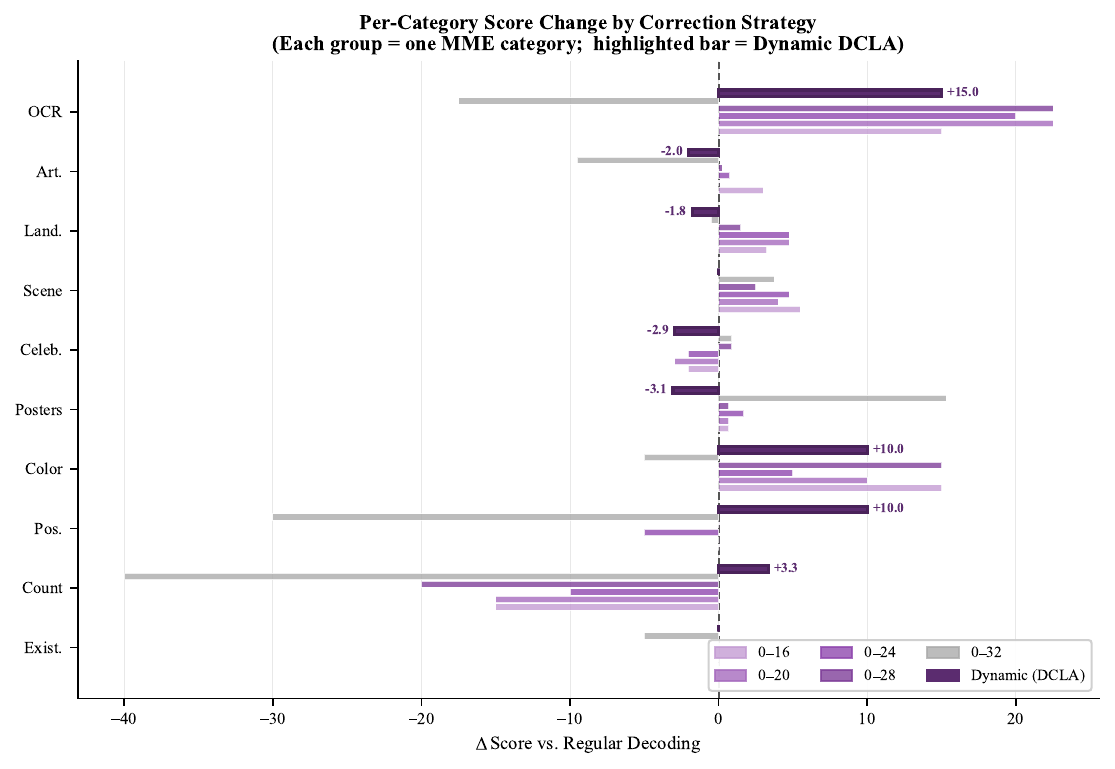}
        \caption{}
        \label{fig:ablation0}
    \end{subfigure}
    \hfill
    \begin{subfigure}[t]{0.56\textwidth}
        \centering
        \includegraphics[width=\textwidth]{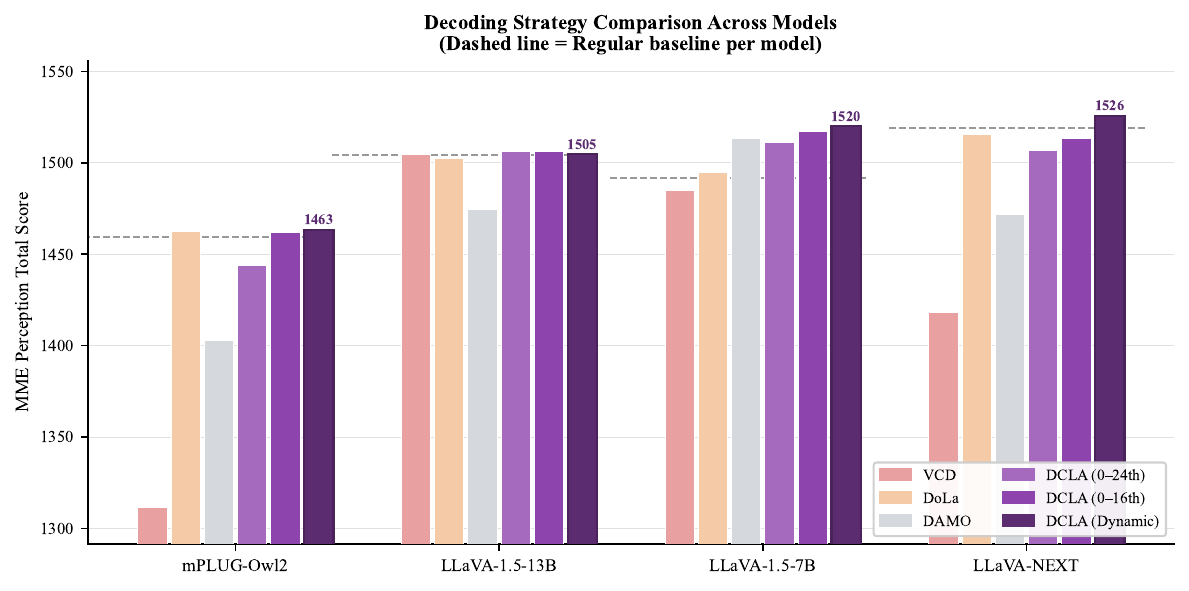}
        \caption{}
        \label{fig:ablation1}
    \end{subfigure}
    \\[0.3em]
    \begin{subfigure}[t]{0.65\textwidth}
        \centering
        \includegraphics[width=\textwidth]{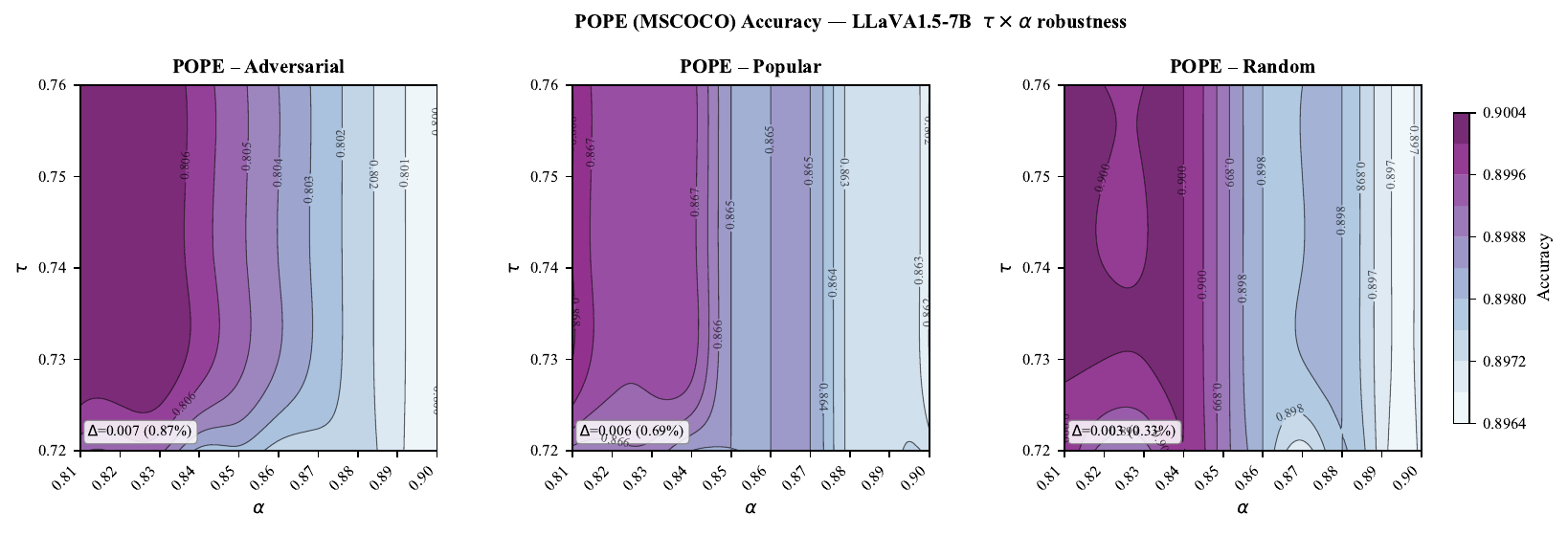}
        \caption{}
        \label{fig:ablation4}
    \end{subfigure}
    \hfill
    \begin{subfigure}[t]{0.32\textwidth}
        \centering
        \includegraphics[width=\textwidth]{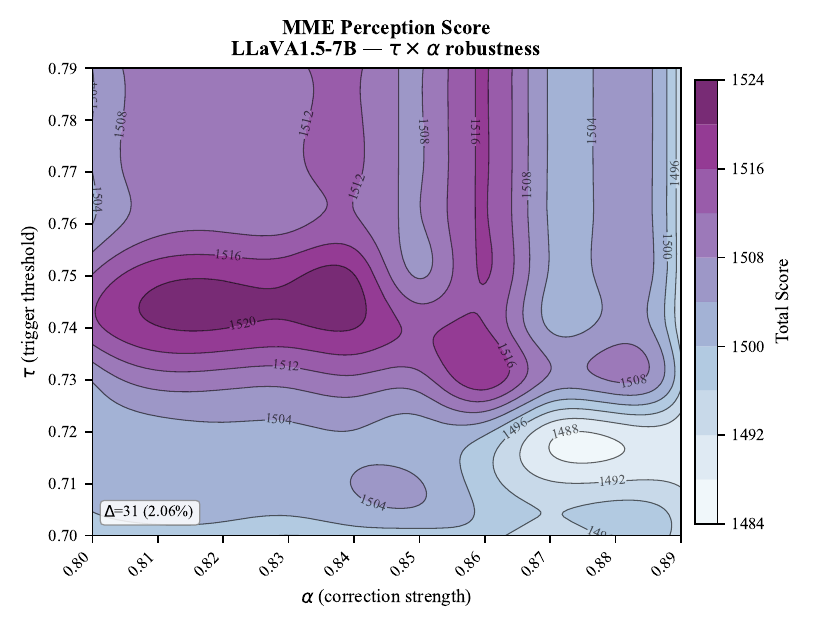}
        \caption{}
        \label{fig:ablation3}
    \end{subfigure}
    \caption{(a) Dynamic DCLA achieves stable per-category gains while fixed ranges show high variance. (b) DCLA (Dynamic) is the only strategy with non-negative $\Delta$ across all four models. (c) POPE accuracy vs.\ $\tau$ and $\alpha$; parallel contours indicate low sensitivity. (d) MME score vs.\ $\tau$ and $\alpha$; peak at $\tau{=}0.74$, $\alpha{=}0.82$.}
    \Description{Four subfigures: (a) bar chart of fixed vs adaptive layer correction on MME; (b) heatmap of dynamic vs fixed correction across models; (c) contour plot of accuracy vs alpha and tau on POPE; (d) contour plot of total score vs alpha and tau on MME.}
\end{figure*}

\subsection{Setup}
\paragraph{Datasets} To comprehensively validate the effectiveness of our DCLA method in mitigating hallucination issues in Large Vision-Language Models (LVLMs), we utilize the MME benchmark \cite{Fu2023MME}. This benchmark comprises 14 diverse tasks, which are categorized under perception and cognition. Additionally, we specifically investigate the effectiveness of DCLA in addressing object hallucinations through benchmark POPE (Polling-based Object Probing Evaluation) \cite{li2023evaluating}, which uses the SEEM-annotated datasets MSCOCO \cite{lin2014microsoft}, A-OKVQA \cite{schwenk2022okvqa} and GQA \cite{hudson2019gqa}. In addition, we further assess the generalization of our method on two real-world VQA benchmarks: VizWiz \cite{gurari2018vizwiz}, which features noisy and ambiguous image inputs, and MM-Vet \cite{yu2023mm}, which provides a comprehensive assessment of multimodal models across six core capabilities. These datasets provide a practical testbed for evaluating the robustness of LVLMs in open-domain settings.

\paragraph{Models and Baselines} We conduct experiments on four recent LVLMs to validate the generalization ability of our method. These include two models with identical architecture but different parameter scales: LLaVA1.5-7b and LLaVA1.5-13b \cite{liu2023visual}, as well as two 7b-scale models with distinct vision-language fusion and pre-training strategies: LLaVA-NEXT \cite{liu2024llavanext} and mPLUG-Owl2 \cite{ye2023mplug}. To further demonstrate the scalability of DCLA beyond the LLaVA family, we additionally evaluate on Qwen2.5-VL-7B-Instruct \cite{bai2025qwen25vl}, Qwen3-VL-8B-Instruct \cite{bai2025qwen3vl}, and InternVL3-8B \cite{zhu2025internvl3}, which employ different vision-language architectures and training recipes. For baseline comparisons, we evaluate DCLA against several representative decoding methods such as regular decoding, VCD \cite{leng2024mitigating}, DoLa \cite{chuang2023dola}, and DAMO \cite{wangdamo}. For DoLa, we fix the mature layer to the 32nd layer and adopt its dynamic candidate-layer selection with an adaptive plausibility constraint; DAMO follows its official configuration. For fair comparison, all decoding strategies are evaluated with temperature set to zero throughout all experiments.

\paragraph{Hyperparameters Setting}
We use only two hyperparameters in DCLA, namely the consistency threshold $\tau$ and the fusion coefficient $\alpha$. Table~\ref{tab:hyperparameters_setting} reports the selected settings for the four main LVLMs. For LLaVA1.5-7B, we further analyze the sensitivity to $\tau$ and $\alpha$ on both POPE and MME. As shown in Figures~\ref{fig:ablation4} and~\ref{fig:ablation3}, performance remains stable across a broad range of values, with the strongest region centered around $\tau=0.74$ and $\alpha=0.82$. Since the detailed grid-search trends are already visualized in the contour plots, we avoid repeating the full numeric grids in additional tables.

\begin{table}[t]
\centering
\caption{DCLA hyperparameter settings for the four main LVLMs.}
\label{tab:hyperparameters_setting}
\resizebox{\columnwidth}{!}{
\begin{tabular}{lcccc}
\toprule
& \textbf{LLaVA1.5-7B} & \textbf{LLaVA-NEXT} & \textbf{LLaVA1.5-13B} & \textbf{mPLUG-Owl2} \\
\midrule
$\alpha$ & 0.82 & 0.96 & 0.86 & 0.90 \\
$\tau$   & 0.74 & 0.93 & 0.76 & 0.95 \\
\bottomrule
\end{tabular}
}
\end{table}

\begin{table*}[t]
\centering
\caption{Evaluation of DCLA and other decoding methods on LLaVA1.5-7b using VizWiz and MM-Vet.}
\label{tab:decoding-comparison}
{\small
\renewcommand{\arraystretch}{0.9}
\setlength{\tabcolsep}{4pt}
\begin{tabular}{l|ccccc|ccccccc}
\toprule
\multirow{2}{*}{\textbf{Decoding}}
& \multicolumn{5}{c|}{\textbf{VizWiz}}
& \multicolumn{7}{c}{\textbf{MM-Vet}} \\
\cmidrule(lr){2-6} \cmidrule(lr){7-13}
& Number & Yes/No & Unans. & Other & \textbf{Overall}
& Rec & OCR & Know & Gen & Spat & Math & \textbf{Total} \\
\midrule
Regular & 47.62 & 78.26 & 74.30 & 38.11 & 50.05 & 36.1 & \textbf{24.5} & 17.5 & 22.2 & 25.7 & \textbf{11.5} & 31.5 \\
VCD                & 42.70 & 77.64 & 72.54 & \textbf{38.12} & 49.47 & 27.7 & 21.5 & 8.3  & 7.6  & 28.1 & 3.8  & 26.1 \\
DoLa               & \textbf{53.33} & \textbf{80.00} & 69.64 & 37.36 & 48.43 & 37.6 & 21.3 & \textbf{21.5} & \textbf{24.6} & 25.6 & 7.7  & 31.8 \\
DAMO               & 48.10 & 78.88 & 71.78 & 36.63 & 48.41 & 35.3 & 21.6 & 21.1 & 21.6 & 24.9 & 7.7  & 31.3 \\
\textbf{DCLA}        & 45.71 & 78.65 & \textbf{77.19} & 37.78 & \textbf{50.62} & \textbf{37.7} & 23.9 & 19.9 & 24.5 & \textbf{29.3} & 7.7 & \textbf{32.1} \\
\bottomrule
\end{tabular}
}
\end{table*}

\subsection{Results}
\paragraph{Results on MME}
To systematically assess the effectiveness of DCLA in mitigating hallucinations, we adopt the perception subset of the MME benchmark, which comprises 10 tasks and has been widely used in recent studies related to hallucinations. Our experiments are conducted on some representative LVLMs. As shown in Table~\ref{tab:mme_main_zeros}, DCLA consistently outperforms almost all baseline decoding methods across the board, achieving total scores of 1520.14 on LLaVA1.5-7b and 1525.73 on LLaVA-NEXT. Notably, DCLA achieves a total score of 1504.82 on the larger-scale model LLaVA1.5-13b. Furthermore, it maintains strong performance on a structurally different architecture, mPLUG-Owl2, with a total score of 1463.40, which surpasses other decoding-based baselines such as DoLa and DAMO.
Existing decoding-based approaches do not guarantee consistent improvements across models: DoLa has no significant improvement over LLaVA1.5-7b, VCD struggles to maintain stable performance across almost all architectures, and DAMO results in performance degradation on LLaVA-NEXT. These findings highlight the superior generation ability and reliability of DCLA in hallucination mitigation for LVLMs.

\paragraph{Results on POPE}
We evaluate DCLA on the SEEM-annotated MSCOCO, A-OKVQA, and GQA datasets from POPE, each with random, popular, and adversarial subsets. As shown in Table~\ref{tab:POPE_results}, DCLA delivers consistent improvements across all three datasets and settings. On MSCOCO, DCLA achieves a 0.93\% accuracy gain under the adversarial setting of LLaVA1.5-7b. On A-OKVQA, DCLA yields an average accuracy improvement of 1.02\% across four models in the adversarial setting. On GQA, DCLA achieves the best performance on LLaVA-NEXT across all settings, with gains of 0.70\% accuracy and 0.84\% F1 under the random setting of mPLUG-Owl2. In contrast, DoLa, VCD, and DAMO exhibit higher instability and even performance regression in some cases.

\paragraph{Results on General-Purpose Multimodal Benchmarks}
As shown in Table~\ref{tab:decoding-comparison}, DCLA achieves strong overall performance on general-purpose multimodal benchmarks. It obtains 50.62\% Overall accuracy on VizWiz, with 77.19\% in the Unanswerable category. On MM-Vet, it records a Total score of 32.1\%, including 37.7\% in Recognition and 29.3\% in Spatial Reasoning. Compared to other decoding methods, DCLA maintains robust and balanced results across all settings, indicating superior generalization.

\begin{table*}[t!]
\centering
\caption{Results of DCLA on Qwen2.5-VL-7B-Instruct, Qwen3-VL-8B-Instruct, and InternVL3-8B across multiple benchmarks. DCLA adopts the selected hyperparameter settings for each model family.}
\label{tab:qwen_results}
{\small
\setlength{\tabcolsep}{4pt}
\begin{tabular}{ll cc cc c cc c}
\toprule
\multirow{2}{*}{\textbf{Model}} & \multirow{2}{*}{\textbf{Decoding}} & \multicolumn{2}{c}{\textbf{POPE}} & \multicolumn{2}{c}{\textbf{HallusionBench}} & \textbf{MME} & \multicolumn{2}{c}{\textbf{MMBench}} & \textbf{MMStar} \\
\cmidrule(lr){3-4} \cmidrule(lr){5-6} \cmidrule(lr){7-7} \cmidrule(lr){8-9} \cmidrule(lr){10-10}
& & F1 & Acc. & aAcc & fAcc & Overall & EN & CN & \\
\midrule
\multirow{4}{*}{Qwen2.5-VL-7B}
& Greedy & 86.38 & 87.70 & 71.82 & 47.69 & 2301.8 & 84.79 & 82.90 & 65.47 \\
& DAMO   & 86.02 & 87.43 & 71.29 & 46.82 & 2291.8 & 84.11 & 79.21 & 64.60 \\
& VCD    & \textbf{87.58} & \textbf{88.62} & 71.61 & 48.55 & 2330.1 & \textbf{85.22} & 83.25 & \textbf{65.73} \\
& \textbf{DCLA} & 87.46 & 88.24 & \textbf{72.13} & \textbf{48.55} & \textbf{2344.4} & 84.71 & \textbf{83.33} & 65.20 \\
\midrule
\multirow{4}{*}{Qwen3-VL-8B}
& Greedy & 89.52 & 89.88 & 71.50 & 48.84 & 2414.1 & \textbf{86.51} & 85.40 & 65.47 \\
& DAMO   & 89.65 & \textbf{90.06} & \textbf{73.08} & \textbf{50.00} & 2398.8 & 86.43 & \textbf{86.00} & 65.53 \\
& VCD    & 89.55 & 89.72 & 69.93 & 48.55 & 2403.9 & 85.74 & 84.97 & 64.93 \\
& \textbf{DCLA} & \textbf{89.67} & 89.74 & 72.34 & 49.42 & \textbf{2429.1} & \textbf{86.51} & 85.74 & \textbf{65.87} \\
\midrule
\multirow{4}{*}{InternVL3-8B}
& Greedy & 90.68 & 90.89 & 50.32 & 41.91 & 2359.8 & 66.15 & 79.81 & 63.93 \\
& DAMO   & 90.15 & 90.60 & 51.94 & \textbf{43.64} & 2368.0 & 66.92 & \textbf{81.19} & 63.93 \\
& VCD    & 90.94 & 91.06 & 47.50 & 38.73 & 2337.8 & 64.95 & 77.23 & \textbf{65.20} \\
& \textbf{DCLA} & \textbf{91.04} & \textbf{91.22} & \textbf{52.03} & 43.35 & \textbf{2370.2} & \textbf{67.53} & 81.01 & 64.13 \\
\bottomrule
\end{tabular}
}
\end{table*}

\paragraph{Generalization to Other Model Families}
To evaluate whether DCLA generalizes beyond the LLaVA family, we further conduct experiments on Qwen2.5-VL-7B-Instruct, Qwen3-VL-8B-Instruct, and InternVL3-8B, which adopt fundamentally different vision-language architectures and training strategies. As shown in Table~\ref{tab:qwen_results}, DCLA achieves the best MME overall score on both Qwen models (2344.4 on Qwen2.5-VL and 2429.1 on Qwen3-VL), outperforming all baselines by a clear margin. On Qwen2.5-VL, DCLA also attains the highest HallusionBench aAcc (72.13\%) and MMBench-CN (83.33\%). On Qwen3-VL, DCLA achieves the best MMStar score (65.87) and ties for the best MMBench-EN (86.51). On InternVL3-8B, DCLA wins 5 out of 8 metrics, including POPE F1 (91.04), POPE Acc (91.22), HallusionBench aAcc (52.03), MME (2370.2), and MMBench-EN (67.53\%), demonstrating clear advantages over both DAMO and VCD. Notably, while DAMO shows strong results on certain benchmarks (e.g., HallusionBench aAcc of 73.08\% on Qwen3-VL, fAcc of 43.64 on InternVL3), it suffers from significant degradation on others such as MME and MMStar. In contrast, DCLA maintains consistently competitive performance across all benchmarks and three model families, confirming its architecture-agnostic generalization ability.

\paragraph{Evaluation on Adaptive Layer Aggregation Correction}
To verify the effectiveness of our dynamic selection mechanism, we conduct a series of ablation studies. In all experiments, aggregation is consistently performed starting from the 1st layer, while representation refinement is applied to each layer from the 0-th up to the $i$-th layer. All other parameters are kept unchanged to ensure a fair and consistent comparison.
As shown in Figure~\ref{fig:ablation0}, the experimental results on the MME dataset using the LLaVA1.5 model indicate that most fixed refinement layer settings achieve reasonably good performance, except for a few extreme values of $i$. In contrast, the dynamic selection mechanism consistently achieves the highest accuracy across all configurations, demonstrating its flexibility and effectiveness in guiding the decoding process. A similar trend can be observed across the other three models, as illustrated in Figure~\ref{fig:ablation1}, further confirming the robustness and generalizability of our approach.

\begin{table}[h]
\centering
\caption{Per-sample inference latency (ms) on a single A800-80GB GPU. Input: 336$\times$336 image + short question; 20 iterations, 3 warmup. Overhead is relative to greedy decoding.}
\vspace{-2mm}
\label{tab:latency}
{\small
\setlength{\tabcolsep}{2pt}
\begin{tabular}{l cc cc cc cc}
\toprule
\multirow{2}{*}{\textbf{Method}} & \multicolumn{2}{c}{\textbf{LLaVA1.5}} & \multicolumn{2}{c}{\textbf{Qwen2.5-VL}} & \multicolumn{2}{c}{\textbf{Qwen3-VL}} & \multicolumn{2}{c}{\textbf{InternVL3}} \\
\cmidrule(lr){2-3} \cmidrule(lr){4-5} \cmidrule(lr){6-7} \cmidrule(lr){8-9}
& ms & Ovhd & ms & Ovhd & ms & Ovhd & ms & Ovhd \\
\midrule
Vanilla  & 111.6 & ---    & 117.1 & ---    & 98.0  & ---    & 116.1 & ---    \\
VCD      & 371.6 & +233\% & 371.3 & +217\% & 227.2 & +132\% & 465.1 & +301\% \\
DAMO     & 138.3 & +24\%  & 429.3 & +267\% & 208.9 & +113\% & 213.6 & +84\%  \\
\textbf{DCLA} & \textbf{138.2} & \textbf{+24\%} & 275.8 & +135\% & \textbf{111.9} & \textbf{+14\%} & \textbf{132.5} & \textbf{+14\%} \\
\bottomrule
\end{tabular}
}
\vspace{-2.3mm}
\end{table}

\paragraph{Inference Efficiency}
As shown in Table~\ref{tab:latency}, we benchmark the per-sample inference latency of each decoding method. On LLaVA1.5-7B, DCLA and DAMO introduce nearly identical overhead (+24\%), while VCD is significantly slower (+233\%). On Qwen3-VL-8B and InternVL3-8B, DCLA introduces only 14\% overhead relative to greedy decoding, substantially lower than DAMO (+113\% / +84\%) and VCD (+132\% / +301\%). On Qwen2.5-VL-7B the overhead is higher (+135\%), which we attribute to a less optimized code path for this architecture. Overall, DCLA offers the best accuracy--efficiency trade-off among all evaluated methods: it achieves the strongest or near-strongest hallucination mitigation while remaining the lightest decoding-time intervention on three out of four model families. Compared to DAMO, which achieves similar or lower accuracy on most benchmarks (Tables~\ref{tab:mme_main_zeros}--\ref{tab:qwen_results}), DCLA reduces the inference overhead by 6--8$\times$ on Qwen3-VL and InternVL3, making it considerably more practical for deployment.

\paragraph{Effect of Correction Strength and Trigger Threshold}
We evaluate the sensitivity of the LLaVA1.5 model to different correction strength values and trigger threshold on the POPE (MSCOCO setting), and record the corresponding accuracy. As shown in Figure~\ref{fig:ablation4}, the model exhibits consistent performance fluctuations across the Random, Popular, and Adversarial subsets as $\alpha$ varies from 0.8 to 0.9, and $\tau$ varies from 0.7 to 0.8, indicating the model's sensitivity to these hyperparameters. Notably, when $\alpha=0.82$ and $\tau=0.74$, the model achieves the highest accuracy on all three subsets.

To further investigate the effectiveness of this parameter setting, we conduct a sensitivity analysis on the MME benchmark. As shown in Figure~\ref{fig:ablation3}, LLaVA1.5 is sensitive to the values of both correction strength and trigger threshold. When $\alpha$ and $\tau$ are set to the same values as those used on the POPE, specifically $\tau=0.74$ and $\alpha = 0.82$, the model achieves the best overall performance on the MME benchmark. This indicates that the selected parameter combination exhibits good cross-dataset generalization on LLaVA1.5.

\section{Limitations}
DCLA also has limitations. This work focuses on hallucination mitigation in image-text multimodal models and does not extend the proposed mechanism to video-language scenarios, where temporal dynamics and cross-frame consistency present new challenges. In addition, DCLA is an inference-only strategy and does not incorporate supervised signals such as reinforcement learning from human feedback or task-specific fine-tuning, which could further enhance its effectiveness. Finally, the method relies entirely on the model's internal representations and does not utilize external retrieval or grounding modules, limiting its ability to correct misinformation acquired during pretraining.

\section{Conclusion}
In this work, we propose Decoding with Inter-layer Consistency via Layer Aggregation (DCLA), a training-free decoding strategy designed to mitigate hallucinations in Large Vision-Language Models (LVLMs). DCLA introduces an explicit cross-layer semantic reference during decoding by aggregating intermediate representations and dynamically selecting and refining the processing layer to enhance semantic stability and suppress hallucinations. Experimental results demonstrate that DCLA achieves consistent improvements on several challenging hallucination evaluation benchmarks, including MME and POPE, with particularly strong gains under adversarial settings. Moreover, DCLA achieves robust improvements on diverse real-world datasets such as VizWiz and MM-Vet, indicating its ability to generalize beyond hallucination mitigation. Notably, DCLA is highly compatible with existing model architectures and can be seamlessly integrated into mainstream multimodal frameworks. Experiments on Qwen2.5-VL, Qwen3-VL, and InternVL3 further confirm that DCLA generalizes effectively across different model families, while introducing only minimal inference overhead.

%%
%% Bibliography


\begin{thebibliography}{78}

%%% the Web address.

\ifx \showCODEN    \undefined \def \showCODEN     #1{\unskip}     \fi
\ifx \showISBNx    \undefined \def \showISBNx     #1{\unskip}     \fi
\ifx \showISBNxiii \undefined \def \showISBNxiii  #1{\unskip}     \fi
\ifx \showISSN     \undefined \def \showISSN      #1{\unskip}     \fi
\ifx \showLCCN     \undefined \def \showLCCN      #1{\unskip}     \fi
\ifx \shownote     \undefined \def \shownote      #1{#1}          \fi
\ifx \showarticletitle \undefined \def \showarticletitle #1{#1}   \fi
\ifx \showURL      \undefined \def \showURL       {\relax}        \fi
% invisible to TeX
\providecommand\bibfield[2]{#2}
\providecommand\bibinfo[2]{#2}
\providecommand\natexlab[1]{#1}
\providecommand\showeprint[2][]{arXiv:#2}

\bibitem[An et~al\mbox{.}(2024)]%
        {an2024agla}
\bibfield{author}{\bibinfo{person}{Wenbin An}, \bibinfo{person}{Feng Tian}, \bibinfo{person}{Sicong Leng}, \bibinfo{person}{Jiahao Nie}, \bibinfo{person}{Haonan Lin}, \bibinfo{person}{QianYing Wang}, \bibinfo{person}{Guang Dai}, \bibinfo{person}{Ping Chen}, {and} \bibinfo{person}{Shijian Lu}.} \bibinfo{year}{2024}\natexlab{}.
\newblock \showarticletitle{Agla: Mitigating object hallucinations in large vision-language models with assembly of global and local attention}.
\newblock \bibinfo{journal}{\emph{arXiv preprint arXiv:2406.12718}} (\bibinfo{year}{2024}).
\newblock


\bibitem[Bai et~al\mbox{.}(2023)]%
        {bai2023qwen}
\bibfield{author}{\bibinfo{person}{Jinze Bai}, \bibinfo{person}{Shuai Bai}, \bibinfo{person}{Yunfei Chu}, \bibinfo{person}{Zeyu Cui}, \bibinfo{person}{Kai Dang}, \bibinfo{person}{Xiaodong Deng}, \bibinfo{person}{Yang Fan}, \bibinfo{person}{Wenbin Ge}, \bibinfo{person}{Yu Han}, \bibinfo{person}{Fei Huang}, {et~al\mbox{.}}} \bibinfo{year}{2023}\natexlab{}.
\newblock \showarticletitle{Qwen technical report}.
\newblock \bibinfo{journal}{\emph{arXiv preprint arXiv:2309.16609}} (\bibinfo{year}{2023}).
\newblock


\bibitem[Bai et~al\mbox{.}(2025a)]%
        {bai2025qwen3vl}
\bibfield{author}{\bibinfo{person}{Shuai Bai}, \bibinfo{person}{Yuxuan Cai}, \bibinfo{person}{Ruizhe Chen}, \bibinfo{person}{Keqin Chen}, \bibinfo{person}{Xionghui Chen}, \bibinfo{person}{Zesen Cheng}, \bibinfo{person}{Lianghao Deng}, \bibinfo{person}{Wei Ding}, \bibinfo{person}{Chang Gao}, \bibinfo{person}{Chunjiang Ge}, {et~al\mbox{.}}} \bibinfo{year}{2025}\natexlab{a}.
\newblock \showarticletitle{Qwen3-VL Technical Report}.
\newblock \bibinfo{journal}{\emph{arXiv preprint arXiv:2511.21631}} (\bibinfo{year}{2025}).
\newblock


\bibitem[Bai et~al\mbox{.}(2025b)]%
        {bai2025qwen25vl}
\bibfield{author}{\bibinfo{person}{Shuai Bai}, \bibinfo{person}{Keqin Chen}, \bibinfo{person}{Xuejing Liu}, \bibinfo{person}{Jialin Wang}, \bibinfo{person}{Wenbin Ge}, \bibinfo{person}{Sibo Song}, \bibinfo{person}{Kai Dang}, \bibinfo{person}{Peng Wang}, \bibinfo{person}{Shijie Wang}, \bibinfo{person}{Jun Tang}, {et~al\mbox{.}}} \bibinfo{year}{2025}\natexlab{b}.
\newblock \showarticletitle{Qwen2.5-VL Technical Report}.
\newblock \bibinfo{journal}{\emph{arXiv preprint arXiv:2502.13923}} (\bibinfo{year}{2025}).
\newblock


\bibitem[Brandon et~al\mbox{.}(2024)]%
        {brandon2024reducing}
\bibfield{author}{\bibinfo{person}{William Brandon}, \bibinfo{person}{Mayank Mishra}, \bibinfo{person}{Aniruddha Nrusimha}, \bibinfo{person}{Rameswar Panda}, {and} \bibinfo{person}{Jonathan Ragan-Kelley}.} \bibinfo{year}{2024}\natexlab{}.
\newblock \showarticletitle{Reducing transformer key-value cache size with cross-layer attention}. In \bibinfo{booktitle}{\emph{The Thirty-eighth Annual Conference on Neural Information Processing Systems}}.
\newblock


\bibitem[Chen et~al\mbox{.}(2025)]%
        {chen2025attribution}
\bibfield{author}{\bibinfo{person}{Qizhou Chen}, \bibinfo{person}{Taolin Zhang}, \bibinfo{person}{Chengyu Wang}, \bibinfo{person}{Xiaofeng He}, \bibinfo{person}{Dakan Wang}, {and} \bibinfo{person}{Tingting Liu}.} \bibinfo{year}{2025}\natexlab{}.
\newblock \showarticletitle{Attribution analysis meets model editing: Advancing knowledge correction in vision language models with visedit}. In \bibinfo{booktitle}{\emph{Proceedings of the AAAI Conference on Artificial Intelligence}}, Vol.~\bibinfo{volume}{39}. \bibinfo{pages}{2168--2176}.
\newblock


\bibitem[Chen et~al\mbox{.}(2024a)]%
        {chen2024multi}
\bibfield{author}{\bibinfo{person}{Xuweiyi Chen}, \bibinfo{person}{Ziqiao Ma}, \bibinfo{person}{Xuejun Zhang}, \bibinfo{person}{Sihan Xu}, \bibinfo{person}{Shengyi Qian}, \bibinfo{person}{Jianing Yang}, \bibinfo{person}{David Fouhey}, {and} \bibinfo{person}{Joyce Chai}.} \bibinfo{year}{2024}\natexlab{a}.
\newblock \showarticletitle{Multi-object hallucination in vision language models}.
\newblock \bibinfo{journal}{\emph{Advances in Neural Information Processing Systems}}  \bibinfo{volume}{37} (\bibinfo{year}{2024}), \bibinfo{pages}{44393--44418}.
\newblock


\bibitem[Chen et~al\mbox{.}(2020)]%
        {chen2020uniter}
\bibfield{author}{\bibinfo{person}{Yen-Chun Chen}, \bibinfo{person}{Linjie Li}, \bibinfo{person}{Licheng Yu}, \bibinfo{person}{Ahmed El~Kholy}, \bibinfo{person}{Faisal Ahmed}, \bibinfo{person}{Zhe Gan}, \bibinfo{person}{Yu Cheng}, {and} \bibinfo{person}{Jingjing Liu}.} \bibinfo{year}{2020}\natexlab{}.
\newblock \showarticletitle{Uniter: Universal image-text representation learning}. In \bibinfo{booktitle}{\emph{European conference on computer vision}}. Springer, \bibinfo{pages}{104--120}.
\newblock


\bibitem[Chen et~al\mbox{.}(2024b)]%
        {chen2024halc}
\bibfield{author}{\bibinfo{person}{Zhaorun Chen}, \bibinfo{person}{Zhuokai Zhao}, \bibinfo{person}{Hongyin Luo}, \bibinfo{person}{Huaxiu Yao}, \bibinfo{person}{Bo Li}, {and} \bibinfo{person}{Jiawei Zhou}.} \bibinfo{year}{2024}\natexlab{b}.
\newblock \showarticletitle{Halc: Object hallucination reduction via adaptive focal-contrast decoding}.
\newblock \bibinfo{journal}{\emph{arXiv preprint arXiv:2403.00425}} (\bibinfo{year}{2024}).
\newblock


\bibitem[Chuang et~al\mbox{.}(2023)]%
        {chuang2023dola}
\bibfield{author}{\bibinfo{person}{Yung-Sung Chuang}, \bibinfo{person}{Yujia Xie}, \bibinfo{person}{Hongyin Luo}, \bibinfo{person}{Yoon Kim}, \bibinfo{person}{James Glass}, {and} \bibinfo{person}{Pengcheng He}.} \bibinfo{year}{2023}\natexlab{}.
\newblock \showarticletitle{Dola: Decoding by contrasting layers improves factuality in large language models}.
\newblock \bibinfo{journal}{\emph{arXiv preprint arXiv:2309.03883}} (\bibinfo{year}{2023}).
\newblock


\bibitem[Dai et~al\mbox{.}(2023)]%
        {dai2023instructblip}
\bibfield{author}{\bibinfo{person}{W Dai}, \bibinfo{person}{J Li}, \bibinfo{person}{D Li}, \bibinfo{person}{AMH Tiong}, \bibinfo{person}{J Zhao}, \bibinfo{person}{W Wang}, \bibinfo{person}{B Li}, \bibinfo{person}{P Fung}, {and} \bibinfo{person}{S Hoi}.} \bibinfo{year}{2023}\natexlab{}.
\newblock \showarticletitle{Instructblip: towards general-purpose vision-language models with instruction tuning. arxiv}.
\newblock \bibinfo{journal}{\emph{Preprint posted online on June}}  \bibinfo{volume}{15} (\bibinfo{year}{2023}), \bibinfo{pages}{2023}.
\newblock


\bibitem[Devlin et~al\mbox{.}(2019)]%
        {devlin2019bert}
\bibfield{author}{\bibinfo{person}{Jacob Devlin}, \bibinfo{person}{Ming-Wei Chang}, \bibinfo{person}{Kenton Lee}, {and} \bibinfo{person}{Kristina Toutanova}.} \bibinfo{year}{2019}\natexlab{}.
\newblock \showarticletitle{Bert: Pre-training of deep bidirectional transformers for language understanding}. In \bibinfo{booktitle}{\emph{Proceedings of the 2019 conference of the North American chapter of the association for computational linguistics: human language technologies, volume 1 (long and short papers)}}. \bibinfo{pages}{4171--4186}.
\newblock


\bibitem[Donahue et~al\mbox{.}(2014)]%
        {donahue2014decaf}
\bibfield{author}{\bibinfo{person}{Jeff Donahue}, \bibinfo{person}{Yangqing Jia}, \bibinfo{person}{Oriol Vinyals}, \bibinfo{person}{Judy Hoffman}, \bibinfo{person}{Ning Zhang}, \bibinfo{person}{Eric Tzeng}, {and} \bibinfo{person}{Trevor Darrell}.} \bibinfo{year}{2014}\natexlab{}.
\newblock \showarticletitle{Decaf: A deep convolutional activation feature for generic visual recognition}. In \bibinfo{booktitle}{\emph{International conference on machine learning}}. PMLR, \bibinfo{pages}{647--655}.
\newblock


\bibitem[Fu et~al\mbox{.}(2023)]%
        {Fu2023MME}
\bibfield{author}{\bibinfo{person}{Chaoyou Fu}, \bibinfo{person}{Peixian Chen}, \bibinfo{person}{Yunhang Shen}, \bibinfo{person}{Yulei Qin}, \bibinfo{person}{Mengdan Zhang}, \bibinfo{person}{Xu Lin}, \bibinfo{person}{Jinrui Yang}, \bibinfo{person}{Xiawu Zheng}, \bibinfo{person}{Ke Li}, \bibinfo{person}{Xing Sun}, \bibinfo{person}{Yunsheng Wu}, {and} \bibinfo{person}{Rongrong Ji}.} \bibinfo{year}{2023}\natexlab{}.
\newblock \showarticletitle{MME: A Comprehensive Evaluation Benchmark for Multimodal Large Language Models}.
\newblock \bibinfo{journal}{\emph{arXiv preprint arXiv:2306.13394}} (\bibinfo{year}{2023}).
\newblock


\bibitem[Guan et~al\mbox{.}(2024)]%
        {guan2024hallusionbench}
\bibfield{author}{\bibinfo{person}{Tianrui Guan}, \bibinfo{person}{Fuxiao Liu}, \bibinfo{person}{Xiyang Wu}, \bibinfo{person}{Ruiqi Xian}, \bibinfo{person}{Zongxia Li}, \bibinfo{person}{Xiaoyu Liu}, \bibinfo{person}{Xijun Wang}, \bibinfo{person}{Lichang Chen}, \bibinfo{person}{Furong Huang}, \bibinfo{person}{Yaser Yacoob}, {et~al\mbox{.}}} \bibinfo{year}{2024}\natexlab{}.
\newblock \showarticletitle{Hallusionbench: an advanced diagnostic suite for entangled language hallucination and visual illusion in large vision-language models}. In \bibinfo{booktitle}{\emph{Proceedings of the IEEE/CVF Conference on Computer Vision and Pattern Recognition}}. \bibinfo{pages}{14375--14385}.
\newblock


\bibitem[Gunjal et~al\mbox{.}(2024)]%
        {gunjal2024detecting}
\bibfield{author}{\bibinfo{person}{Anisha Gunjal}, \bibinfo{person}{Jihan Yin}, {and} \bibinfo{person}{Erhan Bas}.} \bibinfo{year}{2024}\natexlab{}.
\newblock \showarticletitle{Detecting and preventing hallucinations in large vision language models}. In \bibinfo{booktitle}{\emph{Proceedings of the AAAI Conference on Artificial Intelligence}}, Vol.~\bibinfo{volume}{38}. \bibinfo{pages}{18135--18143}.
\newblock


\bibitem[Gurari et~al\mbox{.}(2018)]%
        {gurari2018vizwiz}
\bibfield{author}{\bibinfo{person}{Danna Gurari}, \bibinfo{person}{Qing Li}, \bibinfo{person}{Abigale~J Stangl}, \bibinfo{person}{Anhong Guo}, \bibinfo{person}{Chi Lin}, \bibinfo{person}{Kristen Grauman}, \bibinfo{person}{Jiebo Luo}, {and} \bibinfo{person}{Jeffrey~P Bigham}.} \bibinfo{year}{2018}\natexlab{}.
\newblock \showarticletitle{Vizwiz grand challenge: Answering visual questions from blind people}. In \bibinfo{booktitle}{\emph{Proceedings of the IEEE conference on computer vision and pattern recognition}}. \bibinfo{pages}{3608--3617}.
\newblock


\bibitem[Han et~al\mbox{.}(2022)]%
        {han2022visual}
\bibfield{author}{\bibinfo{person}{Yudong Han}, \bibinfo{person}{Liqiang Nie}, \bibinfo{person}{Jianhua Yin}, \bibinfo{person}{Jianlong Wu}, {and} \bibinfo{person}{Yan Yan}.} \bibinfo{year}{2022}\natexlab{}.
\newblock \showarticletitle{Visual perturbation-aware collaborative learning for overcoming the language prior problem}.
\newblock \bibinfo{journal}{\emph{arXiv preprint arXiv:2207.11850}} (\bibinfo{year}{2022}).
\newblock


\bibitem[Hartsock and Rasool(2024)]%
        {hartsock2024vision}
\bibfield{author}{\bibinfo{person}{Iryna Hartsock} {and} \bibinfo{person}{Ghulam Rasool}.} \bibinfo{year}{2024}\natexlab{}.
\newblock \showarticletitle{Vision-language models for medical report generation and visual question answering: A review}.
\newblock \bibinfo{journal}{\emph{Frontiers in Artificial Intelligence}}  \bibinfo{volume}{7} (\bibinfo{year}{2024}), \bibinfo{pages}{1430984}.
\newblock


\bibitem[Huang et~al\mbox{.}(2024)]%
        {huang2024opera}
\bibfield{author}{\bibinfo{person}{Qidong Huang}, \bibinfo{person}{Xiaoyi Dong}, \bibinfo{person}{Pan Zhang}, \bibinfo{person}{Bin Wang}, \bibinfo{person}{Conghui He}, \bibinfo{person}{Jiaqi Wang}, \bibinfo{person}{Dahua Lin}, \bibinfo{person}{Weiming Zhang}, {and} \bibinfo{person}{Nenghai Yu}.} \bibinfo{year}{2024}\natexlab{}.
\newblock \showarticletitle{Opera: Alleviating hallucination in multi-modal large language models via over-trust penalty and retrospection-allocation}. In \bibinfo{booktitle}{\emph{Proceedings of the IEEE/CVF Conference on Computer Vision and Pattern Recognition}}. \bibinfo{pages}{13418--13427}.
\newblock


\bibitem[Huang et~al\mbox{.}(2020)]%
        {huang2020dianet}
\bibfield{author}{\bibinfo{person}{Zhongzhan Huang}, \bibinfo{person}{Senwei Liang}, \bibinfo{person}{Mingfu Liang}, {and} \bibinfo{person}{Haizhao Yang}.} \bibinfo{year}{2020}\natexlab{}.
\newblock \showarticletitle{Dianet: Dense-and-implicit attention network}. In \bibinfo{booktitle}{\emph{Proceedings of the AAAI conference on artificial intelligence}}, Vol.~\bibinfo{volume}{34}. \bibinfo{pages}{4206--4214}.
\newblock


\bibitem[Hudson and Manning(2019)]%
        {hudson2019gqa}
\bibfield{author}{\bibinfo{person}{Drew~A Hudson} {and} \bibinfo{person}{Christopher~D Manning}.} \bibinfo{year}{2019}\natexlab{}.
\newblock \showarticletitle{Gqa: A new dataset for real-world visual reasoning and compositional question answering}. In \bibinfo{booktitle}{\emph{Proceedings of the IEEE/CVF conference on computer vision and pattern recognition}}. \bibinfo{pages}{6700--6709}.
\newblock


\bibitem[Ji et~al\mbox{.}(2023)]%
        {ji2023survey}
\bibfield{author}{\bibinfo{person}{Ziwei Ji}, \bibinfo{person}{Nayeon Lee}, \bibinfo{person}{Rita Frieske}, \bibinfo{person}{Tiezheng Yu}, \bibinfo{person}{Dan Su}, \bibinfo{person}{Yan Xu}, \bibinfo{person}{Etsuko Ishii}, \bibinfo{person}{Ye~Jin Bang}, \bibinfo{person}{Andrea Madotto}, {and} \bibinfo{person}{Pascale Fung}.} \bibinfo{year}{2023}\natexlab{}.
\newblock \showarticletitle{Survey of hallucination in natural language generation}.
\newblock \bibinfo{journal}{\emph{ACM computing surveys}} \bibinfo{volume}{55}, \bibinfo{number}{12} (\bibinfo{year}{2023}), \bibinfo{pages}{1--38}.
\newblock


\bibitem[Jia et~al\mbox{.}(2021)]%
        {jia2021scaling}
\bibfield{author}{\bibinfo{person}{Chao Jia}, \bibinfo{person}{Yinfei Yang}, \bibinfo{person}{Ye Xia}, \bibinfo{person}{Yi-Ting Chen}, \bibinfo{person}{Zarana Parekh}, \bibinfo{person}{Hieu Pham}, \bibinfo{person}{Quoc Le}, \bibinfo{person}{Yun-Hsuan Sung}, \bibinfo{person}{Zhen Li}, {and} \bibinfo{person}{Tom Duerig}.} \bibinfo{year}{2021}\natexlab{}.
\newblock \showarticletitle{Scaling up visual and vision-language representation learning with noisy text supervision}. In \bibinfo{booktitle}{\emph{International conference on machine learning}}. PMLR, \bibinfo{pages}{4904--4916}.
\newblock


\bibitem[Jiang et~al\mbox{.}(2024)]%
        {jiang2024interpreting}
\bibfield{author}{\bibinfo{person}{Nick Jiang}, \bibinfo{person}{Anish Kachinthaya}, \bibinfo{person}{Suzie Petryk}, {and} \bibinfo{person}{Yossi Gandelsman}.} \bibinfo{year}{2024}\natexlab{}.
\newblock \showarticletitle{Interpreting and editing vision-language representations to mitigate hallucinations}.
\newblock \bibinfo{journal}{\emph{arXiv preprint arXiv:2410.02762}} (\bibinfo{year}{2024}).
\newblock


\bibitem[Kaul et~al\mbox{.}(2024)]%
        {kaul2024throne}
\bibfield{author}{\bibinfo{person}{Prannay Kaul}, \bibinfo{person}{Zhizhong Li}, \bibinfo{person}{Hao Yang}, \bibinfo{person}{Yonatan Dukler}, \bibinfo{person}{Ashwin Swaminathan}, \bibinfo{person}{CJ Taylor}, {and} \bibinfo{person}{Stefano Soatto}.} \bibinfo{year}{2024}\natexlab{}.
\newblock \showarticletitle{Throne: An object-based hallucination benchmark for the free-form generations of large vision-language models}. In \bibinfo{booktitle}{\emph{Proceedings of the IEEE/CVF Conference on Computer Vision and Pattern Recognition}}. \bibinfo{pages}{27228--27238}.
\newblock


\bibitem[Khandelwal et~al\mbox{.}(2024)]%
        {khandelwal2024cross}
\bibfield{author}{\bibinfo{person}{Aditi Khandelwal}, \bibinfo{person}{Harman Singh}, \bibinfo{person}{Hengrui Gu}, \bibinfo{person}{Tianlong Chen}, {and} \bibinfo{person}{Kaixiong Zhou}.} \bibinfo{year}{2024}\natexlab{}.
\newblock \showarticletitle{Cross-Lingual Multi-Hop Knowledge Editing}. In \bibinfo{booktitle}{\emph{Findings of the Association for Computational Linguistics: EMNLP 2024}}. \bibinfo{pages}{11995--12015}.
\newblock


\bibitem[Kim et~al\mbox{.}(2023)]%
        {kim2023exposing}
\bibfield{author}{\bibinfo{person}{Jae~Myung Kim}, \bibinfo{person}{A Koepke}, \bibinfo{person}{Cordelia Schmid}, {and} \bibinfo{person}{Zeynep Akata}.} \bibinfo{year}{2023}\natexlab{}.
\newblock \showarticletitle{Exposing and mitigating spurious correlations for cross-modal retrieval}. In \bibinfo{booktitle}{\emph{Proceedings of the IEEE/CVF conference on computer vision and pattern recognition}}. \bibinfo{pages}{2585--2595}.
\newblock


\bibitem[Leng et~al\mbox{.}(2024)]%
        {leng2024mitigating}
\bibfield{author}{\bibinfo{person}{Sicong Leng}, \bibinfo{person}{Hang Zhang}, \bibinfo{person}{Guanzheng Chen}, \bibinfo{person}{Xin Li}, \bibinfo{person}{Shijian Lu}, \bibinfo{person}{Chunyan Miao}, {and} \bibinfo{person}{Lidong Bing}.} \bibinfo{year}{2024}\natexlab{}.
\newblock \showarticletitle{Mitigating object hallucinations in large vision-language models through visual contrastive decoding}. In \bibinfo{booktitle}{\emph{Proceedings of the IEEE/CVF Conference on Computer Vision and Pattern Recognition}}. \bibinfo{pages}{13872--13882}.
\newblock


\bibitem[Li et~al\mbox{.}(2025a)]%
        {li2025mitigating}
\bibfield{author}{\bibinfo{person}{Shuo Li}, \bibinfo{person}{Jiajun Sun}, \bibinfo{person}{Guodong Zheng}, \bibinfo{person}{Xiaoran Fan}, \bibinfo{person}{Yujiong Shen}, \bibinfo{person}{Yi Lu}, \bibinfo{person}{Zhiheng Xi}, \bibinfo{person}{Yuming Yang}, \bibinfo{person}{Wenming Tan}, \bibinfo{person}{Tao Ji}, {et~al\mbox{.}}} \bibinfo{year}{2025}\natexlab{a}.
\newblock \showarticletitle{Mitigating Object Hallucinations in MLLMs via Multi-Frequency Perturbations}.
\newblock \bibinfo{journal}{\emph{arXiv preprint arXiv:2503.14895}} (\bibinfo{year}{2025}).
\newblock


\bibitem[Li et~al\mbox{.}(2023)]%
        {li2023evaluating}
\bibfield{author}{\bibinfo{person}{Yifan Li}, \bibinfo{person}{Yifan Du}, \bibinfo{person}{Kun Zhou}, \bibinfo{person}{Jinpeng Wang}, \bibinfo{person}{Wayne~Xin Zhao}, {and} \bibinfo{person}{Ji-Rong Wen}.} \bibinfo{year}{2023}\natexlab{}.
\newblock \showarticletitle{Evaluating object hallucination in large vision-language models}.
\newblock \bibinfo{journal}{\emph{arXiv preprint arXiv:2305.10355}} (\bibinfo{year}{2023}).
\newblock


\bibitem[Li et~al\mbox{.}(2025b)]%
        {li2025drafts}
\bibfield{author}{\bibinfo{person}{Yafu Li}, \bibinfo{person}{Zhilin Wang}, \bibinfo{person}{Tingchen Fu}, \bibinfo{person}{Ganqu Cui}, \bibinfo{person}{Sen Yang}, {and} \bibinfo{person}{Yu Cheng}.} \bibinfo{year}{2025}\natexlab{b}.
\newblock \showarticletitle{From Drafts to Answers: Unlocking LLM Potential via Aggregation Fine-Tuning}.
\newblock \bibinfo{journal}{\emph{arXiv preprint arXiv:2501.11877}} (\bibinfo{year}{2025}).
\newblock


\bibitem[Lin et~al\mbox{.}(2021)]%
        {lin2021truthfulqa}
\bibfield{author}{\bibinfo{person}{Stephanie Lin}, \bibinfo{person}{Jacob Hilton}, {and} \bibinfo{person}{Owain Evans}.} \bibinfo{year}{2021}\natexlab{}.
\newblock \showarticletitle{Truthfulqa: Measuring how models mimic human falsehoods}.
\newblock \bibinfo{journal}{\emph{arXiv preprint arXiv:2109.07958}} (\bibinfo{year}{2021}).
\newblock


\bibitem[Lin et~al\mbox{.}(2014)]%
        {lin2014microsoft}
\bibfield{author}{\bibinfo{person}{Tsung-Yi Lin}, \bibinfo{person}{Michael Maire}, \bibinfo{person}{Serge Belongie}, \bibinfo{person}{James Hays}, \bibinfo{person}{Pietro Perona}, \bibinfo{person}{Deva Ramanan}, \bibinfo{person}{Piotr Doll{\'a}r}, {and} \bibinfo{person}{C~Lawrence Zitnick}.} \bibinfo{year}{2014}\natexlab{}.
\newblock \showarticletitle{Microsoft coco: Common objects in context}. In \bibinfo{booktitle}{\emph{Computer vision--ECCV 2014: 13th European conference, zurich, Switzerland, September 6-12, 2014, proceedings, part v 13}}. Springer, \bibinfo{pages}{740--755}.
\newblock


\bibitem[Liu et~al\mbox{.}(2023b)]%
        {liu2023mitigating}
\bibfield{author}{\bibinfo{person}{Fuxiao Liu}, \bibinfo{person}{Kevin Lin}, \bibinfo{person}{Linjie Li}, \bibinfo{person}{Jianfeng Wang}, \bibinfo{person}{Yaser Yacoob}, {and} \bibinfo{person}{Lijuan Wang}.} \bibinfo{year}{2023}\natexlab{b}.
\newblock \showarticletitle{Mitigating hallucination in large multi-modal models via robust instruction tuning}.
\newblock \bibinfo{journal}{\emph{arXiv preprint arXiv:2306.14565}} (\bibinfo{year}{2023}).
\newblock


\bibitem[Liu et~al\mbox{.}(2024a)]%
        {liu2024llavanext}
\bibfield{author}{\bibinfo{person}{Haotian Liu}, \bibinfo{person}{Chunyuan Li}, \bibinfo{person}{Yuheng Li}, \bibinfo{person}{Bo Li}, \bibinfo{person}{Yuanhan Zhang}, \bibinfo{person}{Sheng Shen}, {and} \bibinfo{person}{Yong~Jae Lee}.} \bibinfo{year}{2024}\natexlab{a}.
\newblock \bibinfo{title}{Llavanext: Improved reasoning, ocr, and world knowledge}.
\newblock


\bibitem[Liu et~al\mbox{.}(2023a)]%
        {liu2023visual}
\bibfield{author}{\bibinfo{person}{Haotian Liu}, \bibinfo{person}{Chunyuan Li}, \bibinfo{person}{Qingyang Wu}, {and} \bibinfo{person}{Yong~Jae Lee}.} \bibinfo{year}{2023}\natexlab{a}.
\newblock \showarticletitle{Visual instruction tuning}.
\newblock \bibinfo{journal}{\emph{Advances in neural information processing systems}}  \bibinfo{volume}{36} (\bibinfo{year}{2023}), \bibinfo{pages}{34892--34916}.
\newblock


\bibitem[Liu et~al\mbox{.}(2024b)]%
        {liu2024reducing}
\bibfield{author}{\bibinfo{person}{Sheng Liu}, \bibinfo{person}{Haotian Ye}, \bibinfo{person}{Lei Xing}, {and} \bibinfo{person}{James Zou}.} \bibinfo{year}{2024}\natexlab{b}.
\newblock \showarticletitle{Reducing hallucinations in vision-language models via latent space steering}.
\newblock \bibinfo{journal}{\emph{arXiv preprint arXiv:2410.15778}} (\bibinfo{year}{2024}).
\newblock


\bibitem[Liu et~al\mbox{.}(2019)]%
        {liu2019roberta}
\bibfield{author}{\bibinfo{person}{Yinhan Liu}, \bibinfo{person}{Myle Ott}, \bibinfo{person}{Naman Goyal}, \bibinfo{person}{Jingfei Du}, \bibinfo{person}{Mandar Joshi}, \bibinfo{person}{Danqi Chen}, \bibinfo{person}{Omer Levy}, \bibinfo{person}{Mike Lewis}, \bibinfo{person}{Luke Zettlemoyer}, {and} \bibinfo{person}{Veselin Stoyanov}.} \bibinfo{year}{2019}\natexlab{}.
\newblock \showarticletitle{Roberta: A robustly optimized bert pretraining approach}.
\newblock \bibinfo{journal}{\emph{arXiv preprint arXiv:1907.11692}} (\bibinfo{year}{2019}).
\newblock


\bibitem[Lovenia et~al\mbox{.}(2023)]%
        {lovenia2023negative}
\bibfield{author}{\bibinfo{person}{Holy Lovenia}, \bibinfo{person}{Wenliang Dai}, \bibinfo{person}{Samuel Cahyawijaya}, \bibinfo{person}{Ziwei Ji}, {and} \bibinfo{person}{Pascale Fung}.} \bibinfo{year}{2023}\natexlab{}.
\newblock \showarticletitle{Negative object presence evaluation (nope) to measure object hallucination in vision-language models}.
\newblock \bibinfo{journal}{\emph{arXiv preprint arXiv:2310.05338}} (\bibinfo{year}{2023}).
\newblock


\bibitem[Lu et~al\mbox{.}(2019)]%
        {lu2019vilbert}
\bibfield{author}{\bibinfo{person}{Jiasen Lu}, \bibinfo{person}{Dhruv Batra}, \bibinfo{person}{Devi Parikh}, {and} \bibinfo{person}{Stefan Lee}.} \bibinfo{year}{2019}\natexlab{}.
\newblock \showarticletitle{Vilbert: Pretraining task-agnostic visiolinguistic representations for vision-and-language tasks}.
\newblock \bibinfo{journal}{\emph{Advances in neural information processing systems}}  \bibinfo{volume}{32} (\bibinfo{year}{2019}).
\newblock


\bibitem[Ma et~al\mbox{.}(2024)]%
        {ma2024survey}
\bibfield{author}{\bibinfo{person}{Yueen Ma}, \bibinfo{person}{Zixing Song}, \bibinfo{person}{Yuzheng Zhuang}, \bibinfo{person}{Jianye Hao}, {and} \bibinfo{person}{Irwin King}.} \bibinfo{year}{2024}\natexlab{}.
\newblock \showarticletitle{A survey on vision-language-action models for embodied ai}.
\newblock \bibinfo{journal}{\emph{arXiv preprint arXiv:2405.14093}} (\bibinfo{year}{2024}).
\newblock


\bibitem[Perry et~al\mbox{.}(2025)]%
        {perry2025dynamic}
\bibfield{author}{\bibinfo{person}{Julian Perry}, \bibinfo{person}{Surasakdi Siripong}, {and} \bibinfo{person}{Thanakorn Phonchai}.} \bibinfo{year}{2025}\natexlab{}.
\newblock \showarticletitle{Dynamic Knowledge Integration for Enhanced Vision-Language Reasoning}.
\newblock \bibinfo{journal}{\emph{arXiv preprint arXiv:2501.08597}} (\bibinfo{year}{2025}).
\newblock


\bibitem[Radford et~al\mbox{.}(2021)]%
        {radford2021learning}
\bibfield{author}{\bibinfo{person}{Alec Radford}, \bibinfo{person}{Jong~Wook Kim}, \bibinfo{person}{Chris Hallacy}, \bibinfo{person}{Aditya Ramesh}, \bibinfo{person}{Gabriel Goh}, \bibinfo{person}{Sandhini Agarwal}, \bibinfo{person}{Girish Sastry}, \bibinfo{person}{Amanda Askell}, \bibinfo{person}{Pamela Mishkin}, \bibinfo{person}{Jack Clark}, {et~al\mbox{.}}} \bibinfo{year}{2021}\natexlab{}.
\newblock \showarticletitle{Learning transferable visual models from natural language supervision}. In \bibinfo{booktitle}{\emph{International conference on machine learning}}. PMLR, \bibinfo{pages}{8748--8763}.
\newblock


\bibitem[Rogers et~al\mbox{.}(2021)]%
        {rogers2021primer}
\bibfield{author}{\bibinfo{person}{Anna Rogers}, \bibinfo{person}{Olga Kovaleva}, {and} \bibinfo{person}{Anna Rumshisky}.} \bibinfo{year}{2021}\natexlab{}.
\newblock \showarticletitle{A primer in BERTology: What we know about how BERT works}.
\newblock \bibinfo{journal}{\emph{Transactions of the association for computational linguistics}}  \bibinfo{volume}{8} (\bibinfo{year}{2021}), \bibinfo{pages}{842--866}.
\newblock


\bibitem[Schwenk et~al\mbox{.}(2022)]%
        {schwenk2022okvqa}
\bibfield{author}{\bibinfo{person}{Dustin Schwenk}, \bibinfo{person}{Apoorv Khandelwal}, \bibinfo{person}{Christopher Clark}, \bibinfo{person}{Kenneth Marino}, {and} \bibinfo{person}{Roozbeh Mottaghi}.} \bibinfo{year}{2022}\natexlab{}.
\newblock \showarticletitle{A-okvqa: A benchmark for visual question answering using world knowledge}. In \bibinfo{booktitle}{\emph{European conference on computer vision}}. Springer, \bibinfo{pages}{146--162}.
\newblock


\bibitem[Sun et~al\mbox{.}(2019)]%
        {sun2019videobert}
\bibfield{author}{\bibinfo{person}{Chen Sun}, \bibinfo{person}{Austin Myers}, \bibinfo{person}{Carl Vondrick}, \bibinfo{person}{Kevin Murphy}, {and} \bibinfo{person}{Cordelia Schmid}.} \bibinfo{year}{2019}\natexlab{}.
\newblock \showarticletitle{Videobert: A joint model for video and language representation learning}. In \bibinfo{booktitle}{\emph{Proceedings of the IEEE/CVF international conference on computer vision}}. \bibinfo{pages}{7464--7473}.
\newblock


\bibitem[Tan and Bansal(2019)]%
        {tan2019lxmert}
\bibfield{author}{\bibinfo{person}{Hao Tan} {and} \bibinfo{person}{Mohit Bansal}.} \bibinfo{year}{2019}\natexlab{}.
\newblock \showarticletitle{Lxmert: Learning cross-modality encoder representations from transformers}.
\newblock \bibinfo{journal}{\emph{arXiv preprint arXiv:1908.07490}} (\bibinfo{year}{2019}).
\newblock


\bibitem[Tenney et~al\mbox{.}(2019)]%
        {tenney2019bert}
\bibfield{author}{\bibinfo{person}{Ian Tenney}, \bibinfo{person}{Dipanjan Das}, {and} \bibinfo{person}{Ellie Pavlick}.} \bibinfo{year}{2019}\natexlab{}.
\newblock \showarticletitle{BERT rediscovers the classical NLP pipeline}.
\newblock \bibinfo{journal}{\emph{arXiv preprint arXiv:1905.05950}} (\bibinfo{year}{2019}).
\newblock


\bibitem[Vaswani et~al\mbox{.}(2017)]%
        {vaswani2017attention}
\bibfield{author}{\bibinfo{person}{Ashish Vaswani}, \bibinfo{person}{Noam Shazeer}, \bibinfo{person}{Niki Parmar}, \bibinfo{person}{Jakob Uszkoreit}, \bibinfo{person}{Llion Jones}, \bibinfo{person}{Aidan~N Gomez}, \bibinfo{person}{{\L}ukasz Kaiser}, {and} \bibinfo{person}{Illia Polosukhin}.} \bibinfo{year}{2017}\natexlab{}.
\newblock \showarticletitle{Attention is all you need}.
\newblock \bibinfo{journal}{\emph{Advances in neural information processing systems}}  \bibinfo{volume}{30} (\bibinfo{year}{2017}).
\newblock


\bibitem[Wan et~al\mbox{.}(2024)]%
        {wan2024contrastive}
\bibfield{author}{\bibinfo{person}{David Wan}, \bibinfo{person}{Jaemin Cho}, \bibinfo{person}{Elias Stengel-Eskin}, {and} \bibinfo{person}{Mohit Bansal}.} \bibinfo{year}{2024}\natexlab{}.
\newblock \showarticletitle{Contrastive region guidance: Improving grounding in vision-language models without training}. In \bibinfo{booktitle}{\emph{European Conference on Computer Vision}}. Springer, \bibinfo{pages}{198--215}.
\newblock


\bibitem[Wang et~al\mbox{.}({[n.\,d.]})]%
        {wangdamo}
\bibfield{author}{\bibinfo{person}{Kaishen Wang}, \bibinfo{person}{Hengrui Gu}, \bibinfo{person}{Meijun Gao}, {and} \bibinfo{person}{Kaixiong Zhou}.} \bibinfo{year}{[n.\,d.]}\natexlab{}.
\newblock \showarticletitle{Damo: Decoding by accumulating activations momentum for mitigating hallucinations in vision-language models}. In \bibinfo{booktitle}{\emph{The Thirteenth International Conference on Learning Representations}}.
\newblock


\bibitem[Wang et~al\mbox{.}(2024)]%
        {wang2024mitigating}
\bibfield{author}{\bibinfo{person}{Lei Wang}, \bibinfo{person}{Jiabang He}, \bibinfo{person}{Shenshen Li}, \bibinfo{person}{Ning Liu}, {and} \bibinfo{person}{Ee-Peng Lim}.} \bibinfo{year}{2024}\natexlab{}.
\newblock \showarticletitle{Mitigating fine-grained hallucination by fine-tuning large vision-language models with caption rewrites}. In \bibinfo{booktitle}{\emph{International Conference on Multimedia Modeling}}. Springer, \bibinfo{pages}{32--45}.
\newblock


\bibitem[Woo et~al\mbox{.}(2025)]%
        {woo2025don}
\bibfield{author}{\bibinfo{person}{Sangmin Woo}, \bibinfo{person}{Donguk Kim}, \bibinfo{person}{Jaehyuk Jang}, \bibinfo{person}{Yubin Choi}, {and} \bibinfo{person}{Changick Kim}.} \bibinfo{year}{2025}\natexlab{}.
\newblock \showarticletitle{Don't miss the forest for the trees: Attentional vision calibration for large vision language models}. In \bibinfo{booktitle}{\emph{Findings of the Association for Computational Linguistics: ACL 2025}}. \bibinfo{pages}{1927--1951}.
\newblock


\bibitem[Wu and Tu(2024)]%
        {wu2024layer}
\bibfield{author}{\bibinfo{person}{Haoyi Wu} {and} \bibinfo{person}{Kewei Tu}.} \bibinfo{year}{2024}\natexlab{}.
\newblock \showarticletitle{Layer-condensed kv cache for efficient inference of large language models}. In \bibinfo{booktitle}{\emph{Proceedings of the 62nd Annual Meeting of the Association for Computational Linguistics (Volume 1: Long Papers)}}. \bibinfo{pages}{11175--11188}.
\newblock


\bibitem[Xiao et~al\mbox{.}(2025)]%
        {xiao2025detecting}
\bibfield{author}{\bibinfo{person}{Wenyi Xiao}, \bibinfo{person}{Ziwei Huang}, \bibinfo{person}{Leilei Gan}, \bibinfo{person}{Wanggui He}, \bibinfo{person}{Haoyuan Li}, \bibinfo{person}{Zhelun Yu}, \bibinfo{person}{Fangxun Shu}, \bibinfo{person}{Hao Jiang}, {and} \bibinfo{person}{Linchao Zhu}.} \bibinfo{year}{2025}\natexlab{}.
\newblock \showarticletitle{Detecting and mitigating hallucination in large vision language models via fine-grained ai feedback}. In \bibinfo{booktitle}{\emph{Proceedings of the AAAI Conference on Artificial Intelligence}}, Vol.~\bibinfo{volume}{39}. \bibinfo{pages}{25543--25551}.
\newblock


\bibitem[Xu et~al\mbox{.}(2024)]%
        {xu2024hallucination}
\bibfield{author}{\bibinfo{person}{Ziwei Xu}, \bibinfo{person}{Sanjay Jain}, {and} \bibinfo{person}{Mohan Kankanhalli}.} \bibinfo{year}{2024}\natexlab{}.
\newblock \showarticletitle{Hallucination is inevitable: An innate limitation of large language models}.
\newblock \bibinfo{journal}{\emph{arXiv preprint arXiv:2401.11817}} (\bibinfo{year}{2024}).
\newblock


\bibitem[Yang et~al\mbox{.}({[n.\,d.]})]%
        {yangunderstanding}
\bibfield{author}{\bibinfo{person}{Tianyun Yang}, \bibinfo{person}{Ziniu Li}, \bibinfo{person}{Juan Cao}, {and} \bibinfo{person}{Chang Xu}.} \bibinfo{year}{[n.\,d.]}\natexlab{}.
\newblock \showarticletitle{Understanding and Mitigating Hallucination in Large Vision-Language Models via Modular Attribution and Intervention}. In \bibinfo{booktitle}{\emph{The Thirteenth International Conference on Learning Representations}}.
\newblock


\bibitem[Ye et~al\mbox{.}(2023)]%
        {ye2023mplug}
\bibfield{author}{\bibinfo{person}{Qinghao Ye}, \bibinfo{person}{Haiyang Xu}, \bibinfo{person}{Guohai Xu}, \bibinfo{person}{Jiabo Ye}, \bibinfo{person}{Ming Yan}, \bibinfo{person}{Yiyang Zhou}, \bibinfo{person}{Junyang Wang}, \bibinfo{person}{Anwen Hu}, \bibinfo{person}{Pengcheng Shi}, \bibinfo{person}{Yaya Shi}, {et~al\mbox{.}}} \bibinfo{year}{2023}\natexlab{}.
\newblock \showarticletitle{mplug-owl: Modularization empowers large language models with multimodality}.
\newblock \bibinfo{journal}{\emph{arXiv preprint arXiv:2304.14178}} (\bibinfo{year}{2023}).
\newblock


\bibitem[Ye et~al\mbox{.}(2024)]%
        {ye2024mplug}
\bibfield{author}{\bibinfo{person}{Qinghao Ye}, \bibinfo{person}{Haiyang Xu}, \bibinfo{person}{Jiabo Ye}, \bibinfo{person}{Ming Yan}, \bibinfo{person}{Anwen Hu}, \bibinfo{person}{Haowei Liu}, \bibinfo{person}{Qi Qian}, \bibinfo{person}{Ji Zhang}, {and} \bibinfo{person}{Fei Huang}.} \bibinfo{year}{2024}\natexlab{}.
\newblock \showarticletitle{mplug-owl2: Revolutionizing multi-modal large language model with modality collaboration}. In \bibinfo{booktitle}{\emph{Proceedings of the ieee/cvf conference on computer vision and pattern recognition}}. \bibinfo{pages}{13040--13051}.
\newblock


\bibitem[Yin et~al\mbox{.}(2024)]%
        {yin2024woodpecker}
\bibfield{author}{\bibinfo{person}{Shukang Yin}, \bibinfo{person}{Chaoyou Fu}, \bibinfo{person}{Sirui Zhao}, \bibinfo{person}{Tong Xu}, \bibinfo{person}{Hao Wang}, \bibinfo{person}{Dianbo Sui}, \bibinfo{person}{Yunhang Shen}, \bibinfo{person}{Ke Li}, \bibinfo{person}{Xing Sun}, {and} \bibinfo{person}{Enhong Chen}.} \bibinfo{year}{2024}\natexlab{}.
\newblock \showarticletitle{Woodpecker: Hallucination correction for multimodal large language models}.
\newblock \bibinfo{journal}{\emph{Science China Information Sciences}} \bibinfo{volume}{67}, \bibinfo{number}{12} (\bibinfo{year}{2024}), \bibinfo{pages}{220105}.
\newblock


\bibitem[Yosinski et~al\mbox{.}(2014)]%
        {yosinski2014transferable}
\bibfield{author}{\bibinfo{person}{Jason Yosinski}, \bibinfo{person}{Jeff Clune}, \bibinfo{person}{Yoshua Bengio}, {and} \bibinfo{person}{Hod Lipson}.} \bibinfo{year}{2014}\natexlab{}.
\newblock \showarticletitle{How transferable are features in deep neural networks?}
\newblock \bibinfo{journal}{\emph{Advances in neural information processing systems}}  \bibinfo{volume}{27} (\bibinfo{year}{2014}).
\newblock


\bibitem[Yu et~al\mbox{.}(2018)]%
        {yu2018deep}
\bibfield{author}{\bibinfo{person}{Fisher Yu}, \bibinfo{person}{Dequan Wang}, \bibinfo{person}{Evan Shelhamer}, {and} \bibinfo{person}{Trevor Darrell}.} \bibinfo{year}{2018}\natexlab{}.
\newblock \showarticletitle{Deep layer aggregation}. In \bibinfo{booktitle}{\emph{Proceedings of the IEEE conference on computer vision and pattern recognition}}. \bibinfo{pages}{2403--2412}.
\newblock


\bibitem[Yu et~al\mbox{.}(2023)]%
        {yu2023mm}
\bibfield{author}{\bibinfo{person}{Weihao Yu}, \bibinfo{person}{Zhengyuan Yang}, \bibinfo{person}{Linjie Li}, \bibinfo{person}{Jianfeng Wang}, \bibinfo{person}{Kevin Lin}, \bibinfo{person}{Zicheng Liu}, \bibinfo{person}{Xinchao Wang}, {and} \bibinfo{person}{Lijuan Wang}.} \bibinfo{year}{2023}\natexlab{}.
\newblock \showarticletitle{Mm-vet: Evaluating large multimodal models for integrated capabilities}.
\newblock \bibinfo{journal}{\emph{arXiv preprint arXiv:2308.02490}} (\bibinfo{year}{2023}).
\newblock


\bibitem[Zeng et~al\mbox{.}(2021)]%
        {zeng2021multi}
\bibfield{author}{\bibinfo{person}{Yan Zeng}, \bibinfo{person}{Xinsong Zhang}, {and} \bibinfo{person}{Hang Li}.} \bibinfo{year}{2021}\natexlab{}.
\newblock \showarticletitle{Multi-grained vision language pre-training: Aligning texts with visual concepts}.
\newblock \bibinfo{journal}{\emph{arXiv preprint arXiv:2111.08276}} (\bibinfo{year}{2021}).
\newblock


\bibitem[Zhang et~al\mbox{.}(2026a)]%
        {zhang2026not}
\bibfield{author}{\bibinfo{person}{Dongxu Zhang}, \bibinfo{person}{Hongqiang Lin}, \bibinfo{person}{Yiding Sun}, \bibinfo{person}{Pengyu Wang}, \bibinfo{person}{Qirui Wang}, \bibinfo{person}{Ning Yang}, {and} \bibinfo{person}{Jihua Zhu}.} \bibinfo{year}{2026}\natexlab{a}.
\newblock \showarticletitle{Not all queries need deep thought: Coficot for adaptive coarse-to-fine stateful refinement}. In \bibinfo{booktitle}{\emph{Ann. Conf. Uncertain. Artif. Intell.}}
\newblock


\bibitem[Zhang et~al\mbox{.}(2026b)]%
        {zhang2026pointcot}
\bibfield{author}{\bibinfo{person}{Dongxu Zhang}, \bibinfo{person}{Yiding Sun}, \bibinfo{person}{Pengcheng Li}, \bibinfo{person}{Yumou Liu}, \bibinfo{person}{Hongqiang Lin}, \bibinfo{person}{Haoran Xu}, \bibinfo{person}{Xiaoxuan Mu}, \bibinfo{person}{Liang Lin}, \bibinfo{person}{Wenbiao Yan}, \bibinfo{person}{Ning Yang}, {et~al\mbox{.}}} \bibinfo{year}{2026}\natexlab{b}.
\newblock \showarticletitle{PointCoT: A Multi-modal Benchmark for Explicit 3D Geometric Reasoning}.
\newblock \bibinfo{journal}{\emph{arXiv Prepr. arXiv:2602.23945}} (\bibinfo{year}{2026}).
\newblock


\bibitem[Zhang et~al\mbox{.}(2026c)]%
        {zhang2026chain}
\bibfield{author}{\bibinfo{person}{Dongxu Zhang}, \bibinfo{person}{Yiding Sun}, \bibinfo{person}{Cheng Tan}, \bibinfo{person}{Wenbiao Yan}, \bibinfo{person}{Ning Yang}, \bibinfo{person}{Jihua Zhu}, {and} \bibinfo{person}{Hiajun Zhang}.} \bibinfo{year}{2026}\natexlab{c}.
\newblock \showarticletitle{Chain-of-Thought Compression Should Not Be Blind: V-Skip for Efficient Multimodal Reasoning via Dual-Path Anchoring}. In \bibinfo{booktitle}{\emph{Ann. Meet. Assoc. Comput. Linguist.}}
\newblock


\bibitem[Zhang et~al\mbox{.}(2025)]%
        {zhang2025not}
\bibfield{author}{\bibinfo{person}{Dongxu Zhang}, \bibinfo{person}{Yujun Wu}, \bibinfo{person}{Yiding Sun}, \bibinfo{person}{Jihua Zhu}, \bibinfo{person}{Jinnan Yang}, \bibinfo{person}{Miao Xin}, {and} \bibinfo{person}{Baoliang Tian}.} \bibinfo{year}{2025}\natexlab{}.
\newblock \showarticletitle{Not All Errors Are Created Equal: ASCoT Addresses Late-Stage Fragility in Efficient LLM Reasoning}.
\newblock \bibinfo{journal}{\emph{arXiv Prepr. arXiv:2508.05282}} (\bibinfo{year}{2025}).
\newblock


\bibitem[Zhao et~al\mbox{.}(2021)]%
        {zhao2021recurrence}
\bibfield{author}{\bibinfo{person}{Jingyu Zhao}, \bibinfo{person}{Yanwen Fang}, {and} \bibinfo{person}{Guodong Li}.} \bibinfo{year}{2021}\natexlab{}.
\newblock \showarticletitle{Recurrence along depth: Deep convolutional neural networks with recurrent layer aggregation}.
\newblock \bibinfo{journal}{\emph{Advances in Neural Information Processing Systems}}  \bibinfo{volume}{34} (\bibinfo{year}{2021}), \bibinfo{pages}{10627--10640}.
\newblock


\bibitem[Zhao et~al\mbox{.}(2024)]%
        {zhao2024mitigating}
\bibfield{author}{\bibinfo{person}{Linxi Zhao}, \bibinfo{person}{Yihe Deng}, \bibinfo{person}{Weitong Zhang}, {and} \bibinfo{person}{Quanquan Gu}.} \bibinfo{year}{2024}\natexlab{}.
\newblock \showarticletitle{Mitigating object hallucination in large vision-language models via image-grounded guidance}.
\newblock \bibinfo{journal}{\emph{arXiv preprint arXiv:2402.08680}} (\bibinfo{year}{2024}).
\newblock


\bibitem[Zhou et~al\mbox{.}(2023b)]%
        {zhou2023infmllm}
\bibfield{author}{\bibinfo{person}{Qiang Zhou}, \bibinfo{person}{Zhibin Wang}, \bibinfo{person}{Wei Chu}, \bibinfo{person}{Yinghui Xu}, \bibinfo{person}{Hao Li}, {and} \bibinfo{person}{Yuan Qi}.} \bibinfo{year}{2023}\natexlab{b}.
\newblock \showarticletitle{Infmllm: A unified framework for visual-language tasks}.
\newblock \bibinfo{journal}{\emph{arXiv preprint arXiv:2311.06791}} (\bibinfo{year}{2023}).
\newblock


\bibitem[Zhou et~al\mbox{.}(2024a)]%
        {zhou2024vision}
\bibfield{author}{\bibinfo{person}{Xingcheng Zhou}, \bibinfo{person}{Mingyu Liu}, \bibinfo{person}{Ekim Yurtsever}, \bibinfo{person}{Bare~Luka Zagar}, \bibinfo{person}{Walter Zimmer}, \bibinfo{person}{Hu Cao}, {and} \bibinfo{person}{Alois~C Knoll}.} \bibinfo{year}{2024}\natexlab{a}.
\newblock \showarticletitle{Vision language models in autonomous driving: A survey and outlook}.
\newblock \bibinfo{journal}{\emph{IEEE Transactions on Intelligent Vehicles}} (\bibinfo{year}{2024}).
\newblock


\bibitem[Zhou et~al\mbox{.}(2023a)]%
        {zhou2023analyzing}
\bibfield{author}{\bibinfo{person}{Yiyang Zhou}, \bibinfo{person}{Chenhang Cui}, \bibinfo{person}{Jaehong Yoon}, \bibinfo{person}{Linjun Zhang}, \bibinfo{person}{Zhun Deng}, \bibinfo{person}{Chelsea Finn}, \bibinfo{person}{Mohit Bansal}, {and} \bibinfo{person}{Huaxiu Yao}.} \bibinfo{year}{2023}\natexlab{a}.
\newblock \showarticletitle{Analyzing and mitigating object hallucination in large vision-language models}.
\newblock \bibinfo{journal}{\emph{arXiv preprint arXiv:2310.00754}} (\bibinfo{year}{2023}).
\newblock


\bibitem[Zhou et~al\mbox{.}(2024b)]%
        {zhou2024value}
\bibfield{author}{\bibinfo{person}{Zhanchao Zhou}, \bibinfo{person}{Tianyi Wu}, \bibinfo{person}{Zhiyun Jiang}, {and} \bibinfo{person}{Zhenzhong Lan}.} \bibinfo{year}{2024}\natexlab{b}.
\newblock \showarticletitle{Value Residual Learning For Alleviating Attention Concentration In Transformers}.
\newblock \bibinfo{journal}{\emph{arXiv preprint arXiv:2410.17897}} (\bibinfo{year}{2024}).
\newblock


\bibitem[Zhu et~al\mbox{.}(2023)]%
        {zhu2023minigpt}
\bibfield{author}{\bibinfo{person}{Deyao Zhu}, \bibinfo{person}{Jun Chen}, \bibinfo{person}{Xiaoqian Shen}, \bibinfo{person}{Xiang Li}, {and} \bibinfo{person}{Mohamed Elhoseiny}.} \bibinfo{year}{2023}\natexlab{}.
\newblock \showarticletitle{Minigpt-4: Enhancing vision-language understanding with advanced large language models}.
\newblock \bibinfo{journal}{\emph{arXiv preprint arXiv:2304.10592}} (\bibinfo{year}{2023}).
\newblock


\bibitem[Zhu et~al\mbox{.}(2025)]%
        {zhu2025internvl3}
\bibfield{author}{\bibinfo{person}{Jinguo Zhu}, \bibinfo{person}{Weiyun Chen}, \bibinfo{person}{Zhe Wang}, \bibinfo{person}{Zhaoyang Li}, \bibinfo{person}{Hao Zhao}, \bibinfo{person}{Haoping Liu}, \bibinfo{person}{Wenhai Wang}, \bibinfo{person}{Xizhou Zhu}, \bibinfo{person}{Lewei Lu}, \bibinfo{person}{Yu Qiao}, {and} \bibinfo{person}{Jifeng Dai}.} \bibinfo{year}{2025}\natexlab{}.
\newblock \showarticletitle{InternVL3: Exploring Advanced Training and Test-Time Recipes for Open-Source Multimodal Models}.
\newblock \bibinfo{journal}{\emph{arXiv preprint arXiv:2504.10479}} (\bibinfo{year}{2025}).
\newblock


\bibitem[Zhu et~al\mbox{.}(2024)]%
        {zhu2024ibd}
\bibfield{author}{\bibinfo{person}{Lanyun Zhu}, \bibinfo{person}{Deyi Ji}, \bibinfo{person}{Tianrun Chen}, \bibinfo{person}{Peng Xu}, \bibinfo{person}{Jieping Ye}, {and} \bibinfo{person}{Jun Liu}.} \bibinfo{year}{2024}\natexlab{}.
\newblock \showarticletitle{Ibd: Alleviating hallucinations in large vision-language models via image-biased decoding}.
\newblock \bibinfo{journal}{\emph{arXiv preprint arXiv:2402.18476}} (\bibinfo{year}{2024}).
\newblock


\end{thebibliography}
\end{document}